%% file: paper.tex
\documentclass[10pt, a4paper]{article}
\usepackage[final]{lrec2026} % this is the new style
\input{packages}
\input{commands}

\title{ParliaBench: An Evaluation and Benchmarking Framework for LLM-Generated Parliamentary Speech}

\name{Marios Koniaris\,\orcidlink{0000-0002-0679-5516}, Argyro Tsipi\, \orcidlink{0009-0000-1983-9373
}  Panayiotis Tsanakas\,\orcidlink{0000-0002-8503-5784}} 
\address{National Technical University of Athens\\
School of Electrical and Computer Engineering, Division of Computer Science\\
Iroon Polytechniou 9, Zographou Campus, 15780 Athens, Greece\\
Email: mkoniari@central.ntua.gr, argyrotsipi@gmail.com, panag@cs.ntua.gr}

\abstract{
 \input{0._abstract} 
 \Keywords{Parliamentary Speech Generation,LLM Evaluation,Political Authenticity,Benchmark Evaluation}
}

\begin{document}

\maketitleabstract

\input{1._introduction}

\input{2._related_work} 
\input{3._dataset}
\input{4._evaluation}
\input{5._setup}

\input{6._results}
\input{7._conclusion}
\clearpage

\section*{Availability}
Both datasets (Training and Generated Data) and finetuned models are available under a CC BY License at the ParliaBench collection on \href{https://huggingface.co/collections/argyrotsipi/parliabench}{Hugging Face}. \\
Resources are openly accessible on \href{https://argyrotsipi.github.io/ParliaBench/}{Website} and \href{https://github.com/ArgyroTsipi/ParliaBench}{GitHub}.

\section*{Limitations}
Our evaluation measures linguistic quality and political authenticity but does not assess argument structure or verify factual accuracy against parliamentary records. Our methods rely entirely on automated metrics without human validation, and LLM-as-a-Judge approaches may carry inherent biases.
This work is intended for research and educational purposes, not deployment in actual democratic processes.

\section*{Acknowledgment}
AWS resources were provided by the National Infrastructures for Research and Technology GRNET and funded by the EU Recovery and Resiliency Facility.

\section*{Ethics}
Our work establishes a benchmark resource for evaluating LLM-generated parliamentary speech in research and educational contexts. These resources should only be used as assistance to human experts with consideration of their limitations and biases. The parliamentary data is derived from publicly available UK parliamentary proceedings (ParlaMint corpus) licensed under Creative Commons Attribution 4.0 International, which our derived datasets maintain.
Generated parliamentary speeches must be clearly identified as AI-generated content and not misrepresented as authentic political speeches from actual parliamentarians.

\section{Bibliographical References}\label{sec:reference}

%\bibliographystyle{lrec2026-natbib}
%\bibliography{polyllm} 
\newpage
%\clearpage
\onecolumn
\section*{Appendices}

% % % Switch to one-column format for appendix

\input{appendix_a}
\input{appendix_b}

\input{appendix_c}
\input{appendix_d}
\input{appendix_e}

\end{document}

%% file: packages.tex
\usepackage{xargs}                      % Use more than one optional parameter in a new commands
\usepackage[pdftex,dvipsnames]{xcolor}
\usepackage{tabularx}
\usepackage{hyperref}
\usepackage{bookmark}
\usepackage{xstring}
\usepackage[colorinlistoftodos,prependcaption,textsize=tiny]{todonotes}
\usepackage{array} 
\usepackage{enumitem,booktabs}
\setlist[description]{font=\normalfont\bfseries, leftmargin=!, labelwidth=3cm}
\usepackage{listings}
\usepackage{amsmath}
\usepackage{multicol,multirow}
\usepackage{lipsum}
\usepackage{enumitem}
\usepackage{orcidlink}
\usepackage[many]{tcolorbox}

\lstdefinestyle{styleNoNumbers}{
  basicstyle=\scriptsize\ttfamily,   % Smaller font to ensure fit
  breaklines=true,                  % Enable line breaking
  breakatwhitespace=true,           % Break lines at whitespace
  columns=flexible,                 % Flexible column alignment
  keepspaces=true,                  % Preserve spaces
  showstringspaces=false,           % Don't mark spaces in strings
  xleftmargin=0pt,                  % No extra left margin
  xrightmargin=0pt,                 % No extra right margin
  numbers=none,                     % Remove line numbers
  numberstyle=\tiny\color{gray},    % (Irrelevant with numbers=none)
  keywordstyle=\color{blue},        % Style for keywords
  stringstyle=\color{red},          % Style for strings
  escapeinside={(*@}{@*)},          % Allow LaTeX inside listings
}

\lstdefinestyle{styleTiny}{
  basicstyle=\tiny\ttfamily,         % Even smaller font
  breaklines=true,                  % Enable line breaking
  breakatwhitespace=true,           % Break lines at whitespace
  columns=flexible,                 % Flexible column alignment
  keepspaces=true,                  % Preserve spaces
  showstringspaces=false,           % Don't mark spaces in strings
  xleftmargin=0pt,                  % No extra left margin
  xrightmargin=0pt,                 % No extra right margin
  numbers=left,                     % Line numbers on left
  numberstyle=\tiny\color{gray},    % Small gray line numbers
  keywordstyle=\color{blue},        % Style for keywords
  stringstyle=\color{red},          % Style for strings
  escapeinside={(*@}{@*)},          % Allow LaTeX inside listings
}

\tcbset{
    speechbox/.style={
        colback=gray!5,
        colframe=black!75,
        boxrule=0.5pt,
        arc=2mm,
        left=4pt,
        right=4pt,
        top=4pt,
        bottom=4pt,
        width=0.95\textwidth,
        fonttitle=\bfseries
    },
        templatebox/.style={
        colback=gray!5,
        colframe=black!75,
        boxrule=0.5pt,
        arc=2mm,
        fontupper=\ttfamily\small,
        left=4pt,
        right=4pt,
        top=4pt,
        bottom=4pt
    }
}

%% file: commands.tex
\makeatletter
\newcommand*\iftodonotes{\if@todonotes@disabled\expandafter\@secondoftwo\else\expandafter\@firstoftwo\fi}  
\makeatother

\newcommandx{\finalpass}[2][1=]{\todo[linecolor=red,backgroundcolor=red!50,bordercolor=red,#1]{General check $\rightarrow$ #2}}
\newcommandx{\add}[2][1=]{\todo[color=red,#1]{Add $\rightarrow$ #2}}
\newcommandx{\improve}[2][1=]{\todo[linecolor=OliveGreen,backgroundcolor=OliveGreen!25,bordercolor=OliveGreen,#1]{Improve $\rightarrow$ #2}}
\newcommandx{\change}[2][1=]{\todo[linecolor=blue,backgroundcolor=blue!25,bordercolor=blue,#1]{Change $\rightarrow$ #2}}

\newcommandx{\Finalpass}[2][1=]{\todo[inline,linecolor=red,backgroundcolor=red!50,bordercolor=red,#1]{General check $\rightarrow$ #2}}
\newcommandx{\Add}[2][1=]{\todo[inline,color=red,#1]{Add $\rightarrow$ #2}}
\newcommandx{\Improve}[2][1=]{\todo[inline,linecolor=OliveGreen,backgroundcolor=OliveGreen!25,bordercolor=OliveGreen,#1]{Improve $\rightarrow$ #2}}
\newcommandx{\Change}[2][1=]{\todo[inline,linecolor=blue,backgroundcolor=blue!25,bordercolor=blue,#1]{Change $\rightarrow$ #2}}

%% file: 0._abstract.tex
% !TEX root = paper.tex
Parliamentary speech generation presents specific challenges for large language models beyond standard text generation tasks. Unlike general text generation, parliamentary speeches require not only linguistic quality but also political authenticity and ideological consistency. Current language models lack specialized training for parliamentary contexts, and existing evaluation methods focus on standard NLP metrics rather than political authenticity. To address this, we present ParliaBench, a benchmark for parliamentary speech generation. We constructed a dataset of speeches from UK Parliament to enable systematic model training. We introduce an evaluation framework combining computational metrics with LLM-as-a-judge assessments for measuring generation quality across three dimensions: linguistic quality, semantic coherence, and political authenticity. We propose two novel embedding-based metrics, Political Spectrum Alignment and Party Alignment, to quantify ideological positioning. We fine-tuned five large language models (LLMs), generated 28k speeches, and evaluated them 
using our framework, comparing baseline and fine-tuned models. Results show that fine-tuning produces statistically significant improvements across the majority of metrics and our novel metrics demonstrate strong discriminative power for political dimensions.

%% file: 1._introduction.tex
% !TEX root = paper.tex

\section{Introduction} \label{sec:intro}

Democracy thrives through debate. Democratic parliaments are open forums where elected representatives engage in arguments over policy \cite{hanna021politics}. These debates provide unique insights into political reasoning and ideological positioning. Researchers in political science and computational linguistics increasingly seek to understand and model parliamentary debates. 

Yet generating authentic parliamentary speech presents significant challenges that extend well beyond typical text generation tasks. The complexity becomes apparent when considering specific examples. A Labour MP discussing taxation policy must sound distinctly different from a Conservative counterpart not just in policy position, but most significantly, in their fundamental approach to governance. Large language models have shown promise across political applications, from sentiment analysis \cite{Bestvater2023} and election forecasting \cite{li2024political} to synthetic survey data generation \cite{argyle2023out,bisbee2024synthetic}. Authentic parliamentary speech generation, however, remains challenging.

Evaluation approaches for generated political text fall short for these specialized requirements. Although evaluation methods for text generation tasks have evolved from simple overlap metrics \cite{papineni2002bleu,lin2004rouge} toward more sophisticated embedding-based approaches \cite{zhang2020bertscore,zhao2019moverscore}, they still focus on surface-level similarity rather than political authenticity.

\input{methodology_diagram.tex}

Similarly, domain-specific benchmarks have emerged for legislative summarization \cite{kornilova2019billsum}, opinion alignment \cite{santurkar2023whose}, and other specialized fields \cite{xie2024finben,li2024lexeval}. Parliamentary speech generation, however, lacks the evaluation framework needed to assess ideological consistency and parliamentary conventions simultaneously.

\textbf{Contributions.} We address these limitations by establishing a benchmark resource designed 
specifically for evaluating parliamentary speech generation. First, we develop a curated dataset containing 448k speeches from UK Parliament (Section~\ref{sec:dataset}). Second, we introduce a multi-dimensional evaluation framework that assesses both linguistic quality and political authenticity (Section~\ref{sec:evaluation}). Third, we fine-tune five large language models and generate 28k parliamentary speeches to establish baseline performance (Section~\ref{sec:setup}). Finally, we demonstrate the framework's effectiveness through systematic evaluation (Section~\ref{sec:results}). Figure \ref{fig:methodology_overview} provides an overview of the complete ParliaBench framework and experimental methodology.

%% file: methodology_diagram.tex
% !TEX root = paper.tex
% \begin{figure*}[t]
% \centering
% \resizebox{\textwidth}{!}{%
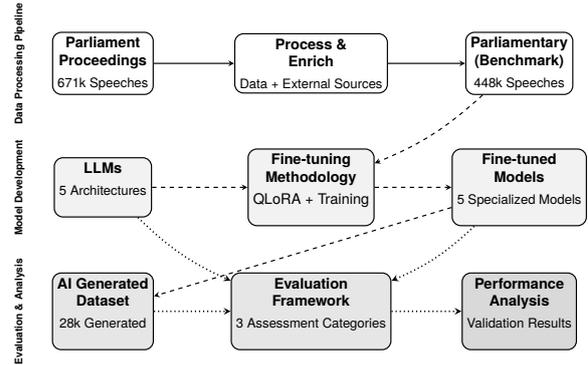
\begin{figure}[t]
\centering
\resizebox{\columnwidth}{!}{
\begin{tikzpicture}[
    box/.style={rectangle, rounded corners=6pt, minimum width=1.8cm, minimum height=1.3cm, align=center, draw=black, thick},
    databox/.style={box, fill=white},
    processbox/.style={box, fill=white},  
    frameworkbox/.style={box, fill=gray!20},
    modelbox/.style={box, fill=gray!10},
    resultbox/.style={box, fill=gray!30},
    arrow/.style={->, thick, >=stealth}
]

% 3x3 Grid Layout
% Row 1: Data Pipeline (y=0)
\node[databox] (input) at (0,0) {
    \textbf{Parliament} \\
    \textbf{Proceedings} \\[0.1cm]
    \small
    % ParlaMint-GB \\
    671k Speeches
    % 2015-2022
};

\node[processbox] (process) at (5,0) {
    \textbf{Process \&} \\
    \textbf{Enrich} \\[0.1cm]
    \small
    Data + External Sources
    % EUROVOC Classification
};

\node[databox] (dataseta) at (10,0) {
    \textbf{Parliamentary} \\
    \textbf{(Benchmark)} \\[0.1cm]
    \small
    448k Speeches
    % Parliamentary \\
    % Debates Benchmark \\
    % 21 EUROVOC Topics
};

% Row 2: Model Pipeline (y=-4)
\node[modelbox] (basemodels) at (0,-3) {
    \textbf{LLMs} \\[0.1cm]
    \small
    5 Architectures \\
    % Llama, Gemma \\
    % Mistral, Qwen2 \\
    % Falcon-H1
};

\node[modelbox] (finetuning) at (5,-3) {
    \textbf{Fine-tuning} \\
    \textbf{Methodology} \\[0.1cm]
    % \small
     QLoRA + Training \\
    % Chat Templates \\
    % Training Pipeline
};

\node[modelbox] (finetuned) at (10,-3) {
    \textbf{Fine-tuned} \\
    \textbf{Models} \\[0.1cm]
    \small
    5 Specialized Models \\
    % Political Discourse \\
    % Specialized
};

% Row 3: Evaluation Pipeline (y=-8)
\node[frameworkbox] (datasetb) at (0,-6) {
    \textbf{AI Generated} \\
    \textbf{Dataset} \\[0.1cm]
    \small
    28k Generated \\
    % 8 Metrics Total \\
    % Multi-dimensional
};

\node[frameworkbox] (framework) at (5,-6) {
    \textbf{Evaluation} \\
    \textbf{Framework} \\[0.1cm]
    \small
    3 Assessment Categories \\
    % Speeches \\
    % 2k per Model
};

% \node[frameworkbox] (framework) at (0,-8) {
%     \textbf{Evaluation} \\
%     \textbf{Framework} \\[0.1cm]
%     \small
%     3 Assessment Categories \\
%     % 8 Metrics Total \\
%     % Multi-dimensional
% };

% \node[databox] (datasetb) at (5,-8) {
%     \textbf{AI Generated} \\
%     \textbf{Dataset} \\[0.1cm]
%     \small
%     10k Generated \\
%     % Speeches \\
%     % 2k per Model
% };

\node[resultbox] (analysis) at (10,-6) {
    \textbf{Performance} \\
    \textbf{Analysis} \\[0.1cm]
    \small
    Validation Results \\
    % Political Authenticity \\
    % Cross-Context Stability
};

% Horizontal arrows (within rows) - different styles per row
\draw[arrow] (input) -- (process);  % Row 1: solid arrows
\draw[arrow] (process) -- (dataseta);
\draw[arrow, dashed] (basemodels) -- (finetuning);  % Row 2: dashed arrows
\draw[arrow, dashed] (finetuning) -- (finetuned);
\draw[arrow, dotted, line width=1.1pt] (datasetb) -- (framework);  % Row 3: dotted arrows
\draw[arrow, dotted, line width=1.1pt] (framework) -- (analysis);

% Vertical connections (between rows)
\draw[arrow, dashed] (finetuned) -- (datasetb);

% Training data connection (dashed)
\draw[arrow, dashed] (dataseta) to [bend left=10] (finetuning);

% Performance Analysis inputs (multiple sources)
%\draw[arrow, dotted, line width=1pt] (framework) -- (datasetb);
%\draw[arrow, dotted, line width=1pt] (datasetb) -- (analysis);
\draw[arrow, dotted, line width=1.1pt] (basemodels) to [bend right=10] (framework);
\draw[arrow, dotted, line width=1.1pt] (finetuned) to [bend left=10] (framework);

% Grid labels - all vertical and left-aligned
%\node[, font=\scriptsize\bfseries, color=black, rotate=90] {Data Processing Pipeline};
%\node[left=0.2cm of basemodels, font=\scriptsize\bfseries, color=black, rotate=90] {Model Development};
%\node[left=0.2cm of datasetb, font=\scriptsize\bfseries, color=black, rotate=90] {Evaluation \& Analysis};
\node[font=\scriptsize\bfseries, color=black, rotate=90] at (-2.0, 0) {\textbf{{Data Processing Pipeline}}};
\node[font=\scriptsize\bfseries, color=black, rotate=90] at (-2.0, -3) {\textbf{{Model Development}}};
\node[font=\scriptsize\bfseries, color=black, rotate=90] at (-2.0, -6) {\textbf{{Evaluation \& Analysis}}};

\end{tikzpicture}
}
\caption{Methodology Overview: Data processing (top row) creates Dataset A from UK Parliament proceedings. Model development (middle row) fine-tunes base architectures using Dataset A. Evaluation pipeline (bottom row) applies the assessment framework to generated speeches (Dataset B) for performance analysis.}
\label{fig:methodology_overview}
\end{figure}
% \caption{Methodology Overview: Data processing (top row) creates Dataset A from UK Parliament proceedings. Model development (middle row) fine-tunes base architectures using Dataset A. Evaluation pipeline (bottom row) applies the assessment framework to generated speeches (Dataset B) for  performance analysis.}
% \label{fig:methodology_overview}
% \end{figure*}

%% file: 2._related_work.tex
% !TEX root = paper.tex
\section{Related Work} \label{sec:related}

Recent advances in large language models (LLMs) have enabled a wide range of applications across political domains. \cite{li2024political} outlines several applications of LLM in political contexts, covering predictive, generative, and simulation-based approaches. The use of LLMs as substitutes for human experts in annotating political texts across multiple languages is explored in \cite{heseltine2024large}, while \cite{gunes2023multiclass} employ LLMs for classifying U.S. Congressional bills. \cite{argyle2023out} investigate LLMs as proxies for specific human subpopulations in social science research and \cite{bisbee2024synthetic} raise concerns about the quality, reliability, and reproducibility of synthetic survey data generated by LLMs.
Agent-based LLMs are utilized as coalition negotiators \cite{moghimifar2024modelling} and as U.S. senators
simulating legislative processes \cite{baker2024simulating}. However, these approaches predominantly emphasize analytical and simulation capabilities rather than authentic speech generation quality.

Traditional text generation evaluation has evolved from reference-based metrics like BLEU \cite{papineni2002bleu} and ROUGE \cite{lin2004rouge} toward embedding based approaches that better capture semantic similarity. BERTScore \cite{zhang2020bertscore} uses contextualized embeddings to compute token-level similarity, while MoverScore \cite{zhao2019moverscore} measures semantic transportation cost using Earth Mover's Distance. For reference-free evaluation, Zhu and Bhat \cite{zhu2020gruen} propose GRUEN, assessing grammaticality and semantic coherence. Domain-specific datasets like BillSum \cite{kornilova2019billsum} for legislative summarization and OpinionQA \cite{santurkar2023whose} for opinion alignment provide targeted evaluation resources, though gaps remain in generative parliamentary speech assessment.

The emergence of LLM-as-a-Judge evaluation \cite{zheng2023judging} offers scalable alternatives for nuanced assessment, achieving over 80\% agreement with human evaluators in complex judgment tasks. \cite{liu2023voices} further validate this approach, demonstrating GPT-4's high alignment with human thematic coding in political analysis. This methodology has been successfully applied across diverse contexts, from general LLM benchmarking through competitive debates \cite{moniri2025evaluating} to long-context reasoning evaluation in parliamentary debates \cite{tiwari2025debatebench}. These approaches demonstrate the viability of automated evaluation for argumentative and political content, though models exhibit documented biases toward Western, educated populations \cite{durmus2024towards} and systematic preferences in political simulations \cite{qi2024representation}. Recent advances in retrieval-augmented generation and chain-of-thought reasoning provide enhanced capabilities for contextually-grounded political text generation, though their application to parliamentary speech evaluation remains underexplored.

Regarding parliamentary speech data, structured corpora like ParlaMint \cite{erjavec2025parlamint} provide multi-lingual parliamentary proceedings. Embedding-based approaches for political analysis, introduced in \cite{rheault2020embeddings}, demonstrate that embeddings can capture ideological positioning in parliamentary text.

Domain-specific evaluation frameworks have emerged across professional fields, including FinBen \cite{xie2024finben} and LexEval \cite{li2024lexeval}. Political science applications have developed specialized benchmarks for election prediction and legislative analysis, yet these focus primarily on classification and analysis tasks. Parliamentary speech generation has attracted recent computational interest, with work exploring European Parliament consensus building \cite{zhang2025eurocon} and political impersonation authenticity \cite{herbold2024}, but evaluation frameworks remain underdeveloped. Existing approaches focus on narrow aspects like style mimicry rather than systematic quality assessment across linguistic and political authenticity dimensions that parliamentary speech generation requires.

Our work addresses these gaps by establishing a benchmark resource specifically designed for parliamentary speech generation. 

%% file: 3._dataset.tex
% !TEX root = paper.tex
\section{ParliaBench Dataset} \label{sec:dataset}

\subsection{Construction \& Processing Pipeline}

We constructed our dataset from the UK subset of the ParlaMint corpus \cite{erjavec2025parlamint}, using the raw version to retain preprocessing control and apply a four-step processing pipeline:

%\begin{enumerate}[label=\textbf{\arabic*.}, leftmargin=15pt, nosep]
\textbf{Step 1: Data Ingestion \& XML Parsing}. Parsed XML files to isolate individual speeches while preserving hierarchical organization.

\textbf{Step 2: Metadata Extraction \& Temporal Alignment}. Enriched speeches with speaker identity, political affiliation, chamber designation, and session dates. As parliamentary speakers frequently change affiliations and roles during their careers, we employed temporal alignment by cross-referencing speech dates with affiliation histories from corpus metadata.

\textbf{Step 3: Content Processing \& Filtering}. Distinguished substantive content from procedural elements, separating parliamentary prompts from speech responses. Filtered procedural noise and non-substantive speeches. For political affiliations, we applied 1000-speech minimum threshold, reducing from 28 original affiliations to 11 to ensure stable model training.

\textbf{Step 4: Thematic Classification}. While ParlaMint uses CAP classification, we selected EuroVoc \cite{eurovoc2025} as the standard classification system for European parliamentary systems. For policy domains with clear semantic correspondence between CAP and EuroVoc taxonomies, we applied direct mapping rules. For semantically complex or ambiguous categories, we employed the methodology provided by \cite{bocchi2024kevlar}. We argue that this approach is particularly well-suited for our dataset because it was specifically designed for legal and governmental texts. For speeches yielding multiple concepts, we selected the highest individual concept score.  (Appendix~\ref{appendix:mapping} documents the hybrid classification strategy)
%(see supplementary material).
%\end{enumerate}

\subsection{Statistics}

The final dataset contains 447,778 speeches from 1,901 unique speakers across 11 political affiliations, with major parties (Conservative: 263,513; Labour: 108,831) dominating representation. The dataset contains approximately 99.94 million words, with speeches averaging 223 words (median: 99 words). This distribution reflects natural variation in parliamentary speeches, from brief procedural statements to extended policy expositions. Temporal coverage captures significant political events including Brexit debates, and the COVID-19 pandemic response, ensuring exposure to diverse political contexts and rhetorical situations. Note: "Bishops", "Crossbench", and "Non-Affiliated" are not political parties in the traditional sense but formal affiliations in Parliament.

\begin{figure}[ht!]
  \centering
  \includegraphics[width=0.97\linewidth]{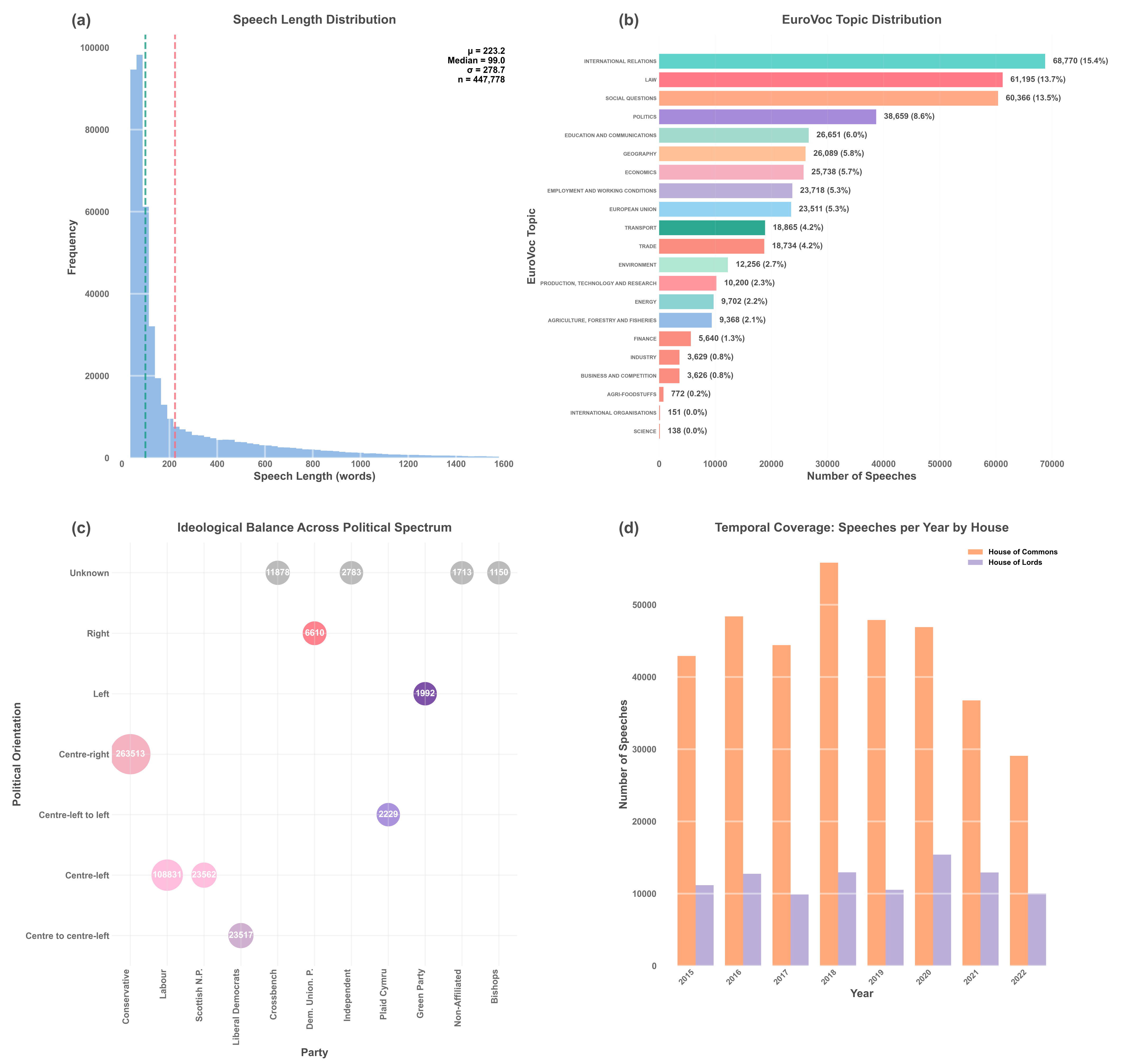}
  \caption{Dataset statistics: (a) Speech length distribution, (b) Topic distribution, (c) Political Orientation distribution, (d) Temporal distribution.}
  \label{fig:statistics}
\end{figure}

Figure \ref{fig:statistics} illustrates key dataset characteristics. Panel (a) reveals a highly right-skewed distribution of speech lengths, indicating that the dataset is dominated by relatively short speeches while containing a smaller number of substantially longer ones. Panel (b) presents topic distribution showing that speeches most frequently address International Relations, Law, and Politics, which together account for over 45\% of the corpus. Panel (c) demonstrates ideological balance across the political spectrum. While most parties by count are in the centre-left to left spectrum, the centre-right to right spectrum produces a larger amount of speeches by 110,000. Panel (d) displays temporal patterns and institutional differences. The House of Commons consistently produces 3-4x more speeches than the House of Lords across 2015-2022.

\subsection{Format and Availability}

The ParliaBench Dataset is distributed as two JSONL datasets:

\textbf{Training Dataset.} An example entry illustrates the dataset structure:
\begin{lstlisting}[style=styleNoNumbers]
{
  "Party": "Labour Party", # Political affiliation  
  "Topic": "Economy",     # Thematic classification
  "Political Orientation": "Left", # Party Political Orientation
  "Section": "Economic Affairs",   # Parliamentary debate section
  "House": "House of Lords",   # Chamber of UK Parliament
  "Speech": "Thank you, Mr. Speaker. I rise to address...", # Complete transcript
  "Prompts": "What measures will your party take..." # Associated prompts
}
\end{lstlisting}

\textbf{Generated Dataset.} We release 27,560 speeches (model outputs) produced during evaluation (\ref{sec:setup}). The generated dataset follows the same format with additional fields: \textit{Model} (architecture identifier), \textit{Type} (baseline/fine-tuned), \textit{Generated Speech} (model output), and \textit{Evaluation Scores} (computed metrics). To ensure a fair comparison, we used the same input prompts across all models and model types, allowing evaluation on identical inputs.

%Both datasets will be available under CC BY License upon publication, with access provided to reviewers upon request.
%Both datasets will be made publicly available under CC BY License upon acceptance.
%Both datasets are available under CC BY License: \href{https://huggingface.co/collections/argyrotsipi/parliabench}{Training Data}, \href{https://huggingface.co/argyrotsipi}{Generated Data}.
Both datasets (Training and Generated Data) and finetuned models are available under a CC BY License at the ParliaBench collection on \href{https://huggingface.co/collections/argyrotsipi/parliabench}{Hugging Face}.

%% file: 4._evaluation.tex
% !TEX root = paper.tex
\section{Evaluation Framework} \label{sec:evaluation}

Parliamentary speech evaluation requires assessment across multiple levels that generic benchmarks cannot capture. Our framework operates on two levels: (i) \textbf{speech evaluation metrics} measuring generation quality across three dimensions, and (ii) \textbf{consistency evaluation metrics} measuring performance reliability across political contexts. This dual-track approach also combines computational metrics with LLM-judge evaluation.

Our framework introduces two novel computational metrics (Political Spectrum  Alignment and Party Alignment, Section~\ref{subsec:novel_metrics}) alongside established metrics from both computational and LLM-judge traditions.

\begin{table}[t]
\small
\centering
\setlength{\tabcolsep}{3pt}
\renewcommand{\arraystretch}{0.92}
\resizebox{\linewidth}{!}{%
\begin{tabular}{l l l l}
    \toprule
    \textbf{Evaluation Level} & \textbf{Assessment Category} & \textbf{Computational Metrics} & \textbf{LLM-Judge Metrics} \\
    \midrule
    \multirow{3}{*}{\textbf{Speech}} 
    & Linguistic Quality & PPL, Dist-N, Self-BLEU & J\_Coh, J\_Conc \\
    \cmidrule(lr){2-4}
    & Semantic Coherence & GRUEN, BERTScore, MoverScore & J\_Rel \\
    \cmidrule(lr){2-4}
    & Political Authenticity & PSA, Party Align & J\_Auth, J\_PolApp, J\_Qual \\
    \midrule
    \textbf{Consistency} 
    & Cross-Context Stability & \multicolumn{2}{c}{Applied to all speech evaluation metrics} \\
    \bottomrule
\end{tabular}%
}
%\captionsetup{justification=raggedright,singlelinecheck=false}
\caption{Multi-Dimensional Evaluation Framework.}
\label{tab:eval_framework}
%\vspace{-0.8em}
\end{table}

Computational metrics provide deterministic assessment through established NLP measures, while LLM-judge metrics capture nuanced qualities requiring contextual understanding. For LLM-judge evaluation, we adapt the methodology from \cite{zheng2023judging}. We employed Flow-Judge-v0.1 as our LLM judge, an LLM specialized in system evaluation tasks. It inherits it's architecture from Phi-3.5-mini instruct \cite{abdin2024}, thus ensuring complete architectural and training data independence from the evaluated models. Our judge rates each speech on a 1-10 scale with written explanations across six parliamentary specific dimensions. We acknowledge this introduces bias through single model judgment. We employ consistent prompt formatting and evaluate speeches sampled across diverse political contexts. Specific prompts for each dimension are provided in Appendix \ref{appendix:llm_judge_prompts}.
%Specific prompts for each dimension are provided in the supplementary material.
% Specific prompts for each dimension are provided in Appendix~\ref{app:llm_judge_prompts}.

Table \ref{tab:eval_framework} shows our evaluation structure. We assess three quality dimensions:  (i) linguistic quality, (ii) semantic coherence, and (iii) political authenticity using both computational and LLM-judge metrics.

\subsection{Speech Evaluation Metrics}

%\begin{enumerate}[label=\textbf{\arabic*.}, leftmargin=15pt, nosep]
\begin{enumerate}[label=\textbf{\arabic*.}, leftmargin=14pt, itemsep=1pt, topsep=2pt, partopsep=0pt, parsep=0pt]
% 1. Linguistic Quality
\item \textbf{Linguistic Quality}
\begin{enumerate}[label=(\alph*), leftmargin=1pt, nosep]

\item \textbf{Computational Metrics}
\begin{enumerate}[label=\roman*., leftmargin=2pt, nosep]
\item \textbf{Perplexity} (PPL)~\cite{jelinek1977perplexity} measures text naturalness. We compute perplexity using GPT-2 as a fixed reference model across all generated speeches, ensuring cross-model comparability. Lower scores indicate more natural-sounding text according to GPT-2's language distribution. (\textbf{$\downarrow$} better)
\item \textbf{Distinct-$n$} (Dist-N)~\cite{li2016diversity} measures lexical diversity through unique bigram proportions (\textbf{$\uparrow$} better), capturing variety in phrase transitions without repetition.
\item \textbf{Self-BLEU}~\cite{zhu2018texygen} measures intra-model diversity by computing BLEU similarity among variants generated under the same prompt. Lower scores indicate greater output variety. (\textbf{$\downarrow$} better)
\end{enumerate}

\item \textbf{LLM-Judge Metrics}
\begin{enumerate}[label=\roman*., leftmargin=2pt, nosep]
\item \textbf{Coherence} (J\_Coh) evaluates logical argument flow and structural organization (\textbf{$\uparrow$} better).
\item \textbf{Conciseness} (J\_Conc) assesses communication efficiency, drawing from debate evaluation criteria~\cite{moniri2025evaluating} (\textbf{$\uparrow$} better).
\end{enumerate}

\end{enumerate}

% 2. Semantic Coherence
\item \textbf{Semantic Coherence}
\begin{enumerate}[label=(\alph*), leftmargin=1pt, nosep]

\item \textbf{Computational Metrics}
\begin{enumerate}[label=\roman*., leftmargin=2pt, nosep]
\item \textbf{GRUEN}~\cite{zhu2020gruen} evaluates linguistic quality through grammatical correctness and semantic coherence (\textbf{$\uparrow$} better), aggregating four quality dimensions: grammaticality, non-redundancy, focus, and structural coherence.
\item \textbf{BERTScore}~\cite{zhang2020bertscore} measures semantic similarity using RoBERTa-large embeddings (\textbf{$\uparrow$} better), reporting F1-score to balance precision and recall.
\item \textbf{MoverScore}~\cite{zhao2019moverscore} quantifies semantic distance via Earth Mover's Distance (\textbf{$\uparrow$} better).
\item[] \textit{Note:} For both BERTScore and MoverScore, generated speeches are compared against the top-5 human speeches from the training set matching the same context.
\end{enumerate}

\item \textbf{LLM-Judge Metrics}
\begin{enumerate}[label=\roman*., leftmargin=2pt, nosep]
\item \textbf{Relevance} (J\_Rel) evaluates whether responses address given prompts (\textbf{$\uparrow$} better).
\end{enumerate}

\end{enumerate}

% 3. Political Authenticity
\item \textbf{Political Authenticity}
\begin{enumerate}[label=(\alph*), leftmargin=1pt, nosep]
\item \textbf{Computational Metrics}
\begin{enumerate}[label=\roman*., leftmargin=2pt, nosep]
\item We employ two novel embedding-based metrics for political authenticity assessment: Political Spectrum Alignment (PSA) and Party Alignment (Party Align). Detailed calculation methodology provided in Section~\ref{subsec:novel_metrics}.
\end{enumerate}

\item \textbf{LLM-Judge Metrics}
\begin{enumerate}[label=\roman*., leftmargin=2pt, nosep]
\item \textbf{Authenticity} (J\_Auth) assesses whether content reflects genuine political speech (\textbf{$\uparrow$} better).
\item \textbf{Political Appropriateness} (J\_PolApp) evaluates whether tone is suitable for political speech (\textbf{$\uparrow$} better).
\item \textbf{Overall Quality} (J\_Qual) assesses sophistication, persuasiveness, and communicative effectiveness (\textbf{$\uparrow$} better).
\end{enumerate}

\end{enumerate}

\end{enumerate}

 \subsubsection{Novel Political Authenticity Metrics} \label{subsec:novel_metrics}

\textbf{Political Spectrum Alignment} (PSA) evaluates ideological positioning on the left-right spectrum. We adapt semantic embedding approaches \cite{rheault2020embeddings} for LLM-generated speech evaluation, drawing on 
the Left-Right (RILE) scale methodology \cite{volkens2013understanding,budge2013standard}. Our metric employs a two-stage approach combining semantic similarity with ideological distance. We create reference embeddings by grouping parliamentary speeches by political orientations (Far-left through Far-right, including intermediate positions) and computing centroid embeddings using sentence transformers. Orientations map to numerical values where Far-left = -6, Centre = 0, Far-right = +6. 

We first identify the closest matching political orientation:

\begin{equation}
po^* = \text{argmax}_{po \in \mathcal{PO}} \text{sim}(s, c_{po})
\label{eq:political_orientation}
\end{equation}

\noindent where $po^*$ is the closest matching orientation, $\mathcal{PO}$ 
the set of all political orientations, $c_{po}$ the orientation centroid, and $\text{sim}(s, c_{po})$ the cosine similarity between generated speech $s$ and centroid.

The PSA score combines semantic similarity with orientation distance:

% \begin{equation}
% \text{PSA}(s, po_e) = \text{sim}(s, c_{po^*}) \times \max\left(0, 1 - \frac{|\phi(po_e) - \phi(po^*)|}{12}\right)
% \label{eq:psa}
% \end{equation}
\begin{equation}
\text{PSA}(s, po_e) = \text{sim}(s, c_{po^*}) \cdot \max\left(0, 1 - \frac{\Delta_\phi}{12}\right)
\label{eq:psa}
\end{equation}
\noindent where $\Delta_\phi = |\phi(po_e) - \phi(po^*)|$, $po_e$ is the expected orientation, $\phi(po)$ maps orientations to numerical values, and $\text{sim}(s, c_{po^*})$ is the cosine similarity to the matched centroid. The maximum distance of 12 (from Far-left to Far-right) normalizes to 0-1 scale. Perfect ideological alignment approaches 1, while misalignment approaches 0.

%\noindent where $po_e$ is the expected orientation, $\phi(po)$ maps orientations to numerical values, and $\text{sim}(s, c_{po^*})$ is the cosine similarity to the matched centroid. The maximum distance of 12 (from Far-left to Far-right) normalizes to 0-1 scale. Perfect ideological alignment approaches 1, while misalignment approaches 0.
% where 1 indicates perfect ideological alignment and 0 represents complete misalignment.
\noindent Parties with unknown orientation are excluded from PSA analysis as their ideological position cannot be reliably mapped to the left-right spectrum.

\noindent \textbf{Party Alignment} (Party Align) applies the same embedding methodology to party-specific alignment, using party affiliation rather than political orientation for centroid construction. The alignment score measures cosine similarity between generated speech and expected party centroid:

\begin{equation}
\text{Party Align}(s, p) = \text{sim}(s, c_p)
\label{eq:party_align}
\end{equation}

\noindent where $c_p$ is the party-specific centroid and $\text{sim}(s, c_p)$ the cosine similarity between speech $s$ and centroid (0-1 scale). Higher scores indicate stronger alignment with party-specific language.

\subsection{Consistency Evaluation Metrics}

\textbf{Cross-Context Stability} measures performance consistency using coefficient of variation, with higher scores indicating more consistent performance across political contexts. This meta-evaluation applies to all speech evaluation metrics, providing diagnostic insight into model reliability. For cross-metric comparison, all metrics, Computational \& LLM-Judge, are normalized to 0-1 scale.

We examine consistency across three dimensions: political parties, topic domains, and political orientations. The stability calculation quantifies performance variability:
\begin{equation}
\text{Stability} = \frac{100}{1 + CV}
\label{eq:stability}
\end{equation}
where $CV = \frac{\sigma}{\mu}$ represents the ratio of standard deviation to mean across context groups.

%% file: 5._setup.tex
\section{Experimental Setup} \label{sec:setup}

\subsection{Model Architecture Selection}

We selected five language models representing distinct architectural approaches to establish baseline performance for parliamentary speech generation: 
\textbf{Mistral 7B v0.3}~\cite{jiang2023mistral} uses Grouped Query Attention and Sliding Window Attention for efficient long-context processing.
\textbf{Llama 3.1 8B}~\cite{dubey2024llama} features 128k-token context window and expanded vocabulary.
\textbf{Gemma 2 9B}~\cite{team2024gemma} employs alternating local/global attention across 42 layers with logit soft-capping.
\textbf{Qwen2 7B}~\cite{team2024qwen2} is a multilingual model with enhanced reasoning capabilities.
\textbf{YI 6B}~\cite{ai2025yiopenfoundationmodels} emphasizes strong reasoning and coding performance.

\subsection{Fine-Tuning and Training Configuration}

We employed Quantized Low-Rank Adaptation (QLoRA) \cite{QLORA_2023} for parameter-efficient fine-tuning. ($r=16$, $\alpha=16$, $2$ epochs). Model-specific chat templates structure training inputs with political metadata (party affiliation, topic classification, orientation, section, and house). Training used SFTTrainer from TRL with 80\%-20\% train-test splits and automated checkpointing. Fine-tuned models were saved with adapter weights for subsequent evaluation, ensuring consistent model states across experiments.

\subsection{Evaluation Dataset Generation}

We planned to generate 30,000 speeches (3,000 per model-type combination across 10 models) using stratified sampling from the held-out test set. To maintain consistency with training conditions, our prompt distribution matches the ParliaBench Dataset structure: 90\% generic instruction prompts formatted with political context (party, topic, orientation, section, house) and 10\% specific parliamentary questions from the test set. 

Generation employed nucleus sampling ($temperature=0.7$, $top\-p=0.85$, $repetition_penalty=1.2$). Generated speeches underwent validation for template leakage, encoding corruption, semantic relevance, and length constraints. Invalid outputs were automatically regenerated (max 3 attempts). Baseline models exhibited higher failure rates, suggesting fine-tuning improved output quality. To ensure fair cross-model comparison, we retained only speeches successfully generated by all 10 model-type combinations, yielding 29,220 speeches. 

Generated speeches were then evaluated using our LLM-judge framework. Speeches with incomplete dimension ratings were excluded, resulting in 27,560 fully evaluated speeches, across all six dimensions, for all models. All subsequent results use the 27,560 fully evaluated speeches, which maintain balanced representation across political affiliations and topics. Complete implementation details, including QLoRA configuration, training parameters, chat templates for all architectures, and speech generation validation methodology, are provided in the Appendix \ref{appendix:setup}.
%supplementary material.

%% file: 6._results.tex
% !TEX root = paper.tex
\section{Results and Analysis} \label{sec:results}

We evaluated 27,560 generated speeches using our evaluation framework. This section presents fine-tuning effectiveness and performance patterns across political parties, topic domains, and ideological orientations. Representative examples of generated speeches are provided in Appendix~\ref{appendix:speeches}.
%Representative examples of generated speeches are provided in the supplementary material.

\subsection{Overview and Fine-Tuning Impact}

 \input{tables/performance_table_with_significance2.tex}

Table \ref{tab:performance_table} presents metric results organized by our framework assessment categories. Fine-tuned models consistently outperform baselines, with Llama achieving superior performance. %(GRUEN 0.534±0.018, BERTScore 0.759±0.012). 
%Fine-tuned models demonstrate substantially reduced variance compared to baselines, indicating more consistent generation across diverse political contexts. 
Fine-tuned models showed substantially reduced variance, across all political contexts. Extended context windows (128k tokens) and larger vocabularies contribute to architectural advantages.

Pairwise t-tests confirm statistical significance of fine-tuning effects (45 out of 70 comparisons). %Of 70 comparisons (5 models × 14 metrics), 45 remained significant after Bonferroni correction ($\alpha$=0.05, 64.3\%).
Model architectures exhibited differential responsiveness: \textit{YI} and \textit{Llama} achieved notable improvements (11/14 metrics, 79\%), while others showed more selective gains (improvements marked with $^{*}$ in Table \ref{tab:performance_table}).

Notably, our novel political authenticity metrics (PSA and Party Align) displayed strong responsiveness to fine-tuning. All five models significantly improved PSA ($p<0.001$), with effect sizes ranging from small to very large (d=0.141-1.045). Party Align showed similar patterns (4 of 5 models improved, d=0.099-1.221). These substantial effects validate that our embedding based metrics capture critical political authenticity dimensions unavailable to conventional evaluation. 
%For complete t-test results including effect sizes, see the supplementary material.% 
For complete t-test results including effect sizes, see Table~\ref{tab:ttest_significant_results} in Appendix~\ref{app:statistical_tests}.

\begin{figure}[ht]
  \centering
  \includegraphics[width=0.9\columnwidth,keepaspectratio]{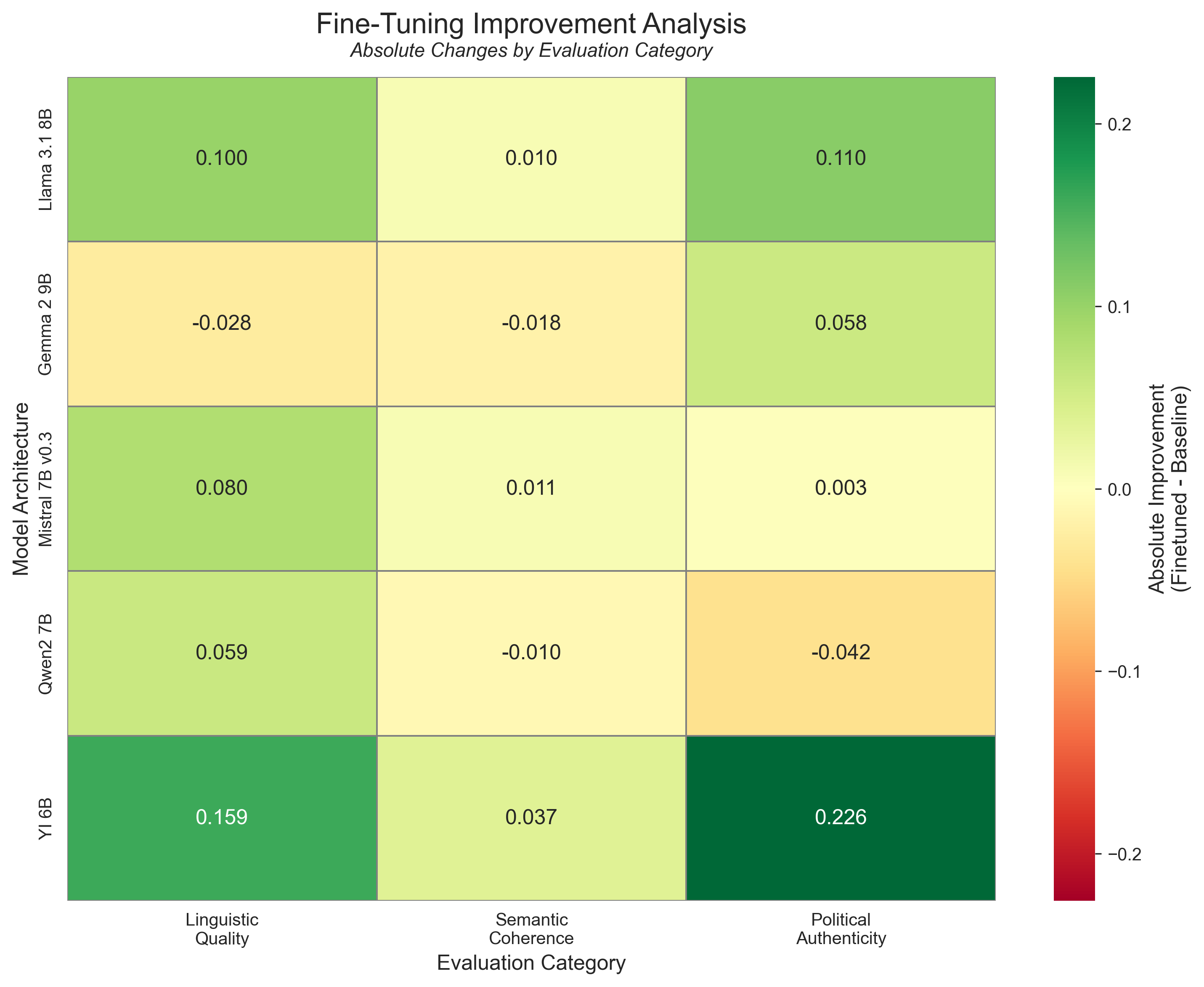}
  \caption{Absolute performance changes (finetuned vs. baseline) across evaluation categories. All metrics normalized to comparable scales.} %Green indicates improvement, red indicates decline.}
  \label{fig:finetuning_improvement_quality}
\end{figure}

Figure \ref{fig:finetuning_improvement_quality} shows fine-tuning impact across evaluation categories. \textit{YI} achieved the strongest improvements across all dimensions %(Linguistic: +0.159, Semantic: +0.037, Political: +0.226)
, while \textit{Llama} had consistent gains. \textit{Gemma2} and \textit{Qwen2} exhibited quality trade-offs, with improvements in one category accompanied by declines in others, suggesting architectural differences in how models balance competing objectives during fine-tuning. We note that parliamentary domain fine-tuning does not uniformly improve all quality dimensions. Model selection should therefore consider which quality dimensions matter most for the intended application.

Cross-context stability analysis (Eq.\ref{eq:stability}) revealed fine-tuned models maintained consistent performance across political contexts (composite stability 91.4-96.2). Mistral achieved highest consistency (96.2) despite trade-offs in absolute performance, while Llama (95.1) balanced strong performance with stability. 
Detailed scores in Table~\ref{tab:stability_breakdown}, Appendix~\ref{app:statistical_tests}.
%(Details in the supplementary material.)

\subsection{Political Context Results}

\begin{figure}[ht]
  \centering
  \includegraphics[width=0.9\columnwidth,keepaspectratio]{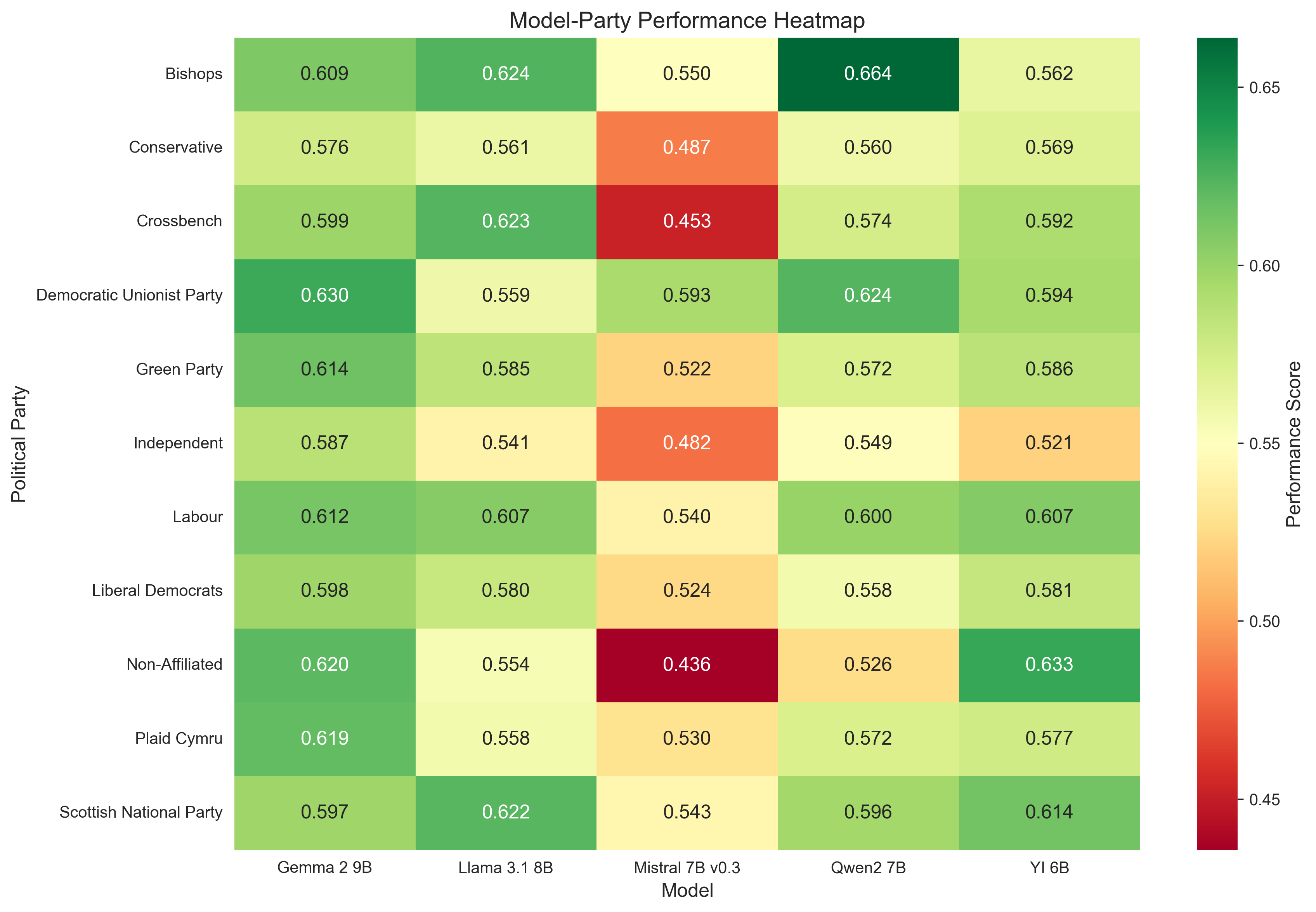}
  \caption{Party alignment scores (0-1 scale) for finetuned models across UK parliamentary parties. Color intensity indicates performance level}% (green = high, red = low).}
  %\caption{Party alignment scores (0-1 scale) for finetuned models.}
  \label{fig:party_performance}
\end{figure}

\textbf{Party alignment Patterns}. Party alignment performance varied substantially across models (Figure \ref{fig:party_performance}). Major parties (Conservative, Labour) achieved stable performance across models, benefiting from substantial training data (58.9\%, 24.3\%). Minor parties exhibited greater variability. \textit{Mistral} struggled with heterogeneous groups (Non-Affiliated: 0.436), while \textit{Qwen} excelled with ideologically coherent minorities (Bishops: 0.664). \textit{YI} demonstrated robust cross-party performance (0.614-0.633). 
Detailed scores in Table~\ref{tab:party_performance}, Appendix~\ref{app:statistical_tests}.
% Detailed scores in the supplementary material.

Both new political authenticity metrics (PSA and Party Align) successfully discriminate their target political dimensions. Party Align distinguishes parties while PSA distinguishes orientations (both $p<0.001$). Our analysis reveals that Party Align performance depends primarily on data abundance and ideological coherence rather than party size alone. Models successfully learn party-specific language patterns when training data provides clear stylistic signals, indicating targeted data collection for under-represented parties could improve coverage.

\begin{figure}[ht]
  \centering
  \includegraphics[width=0.9\columnwidth,keepaspectratio]{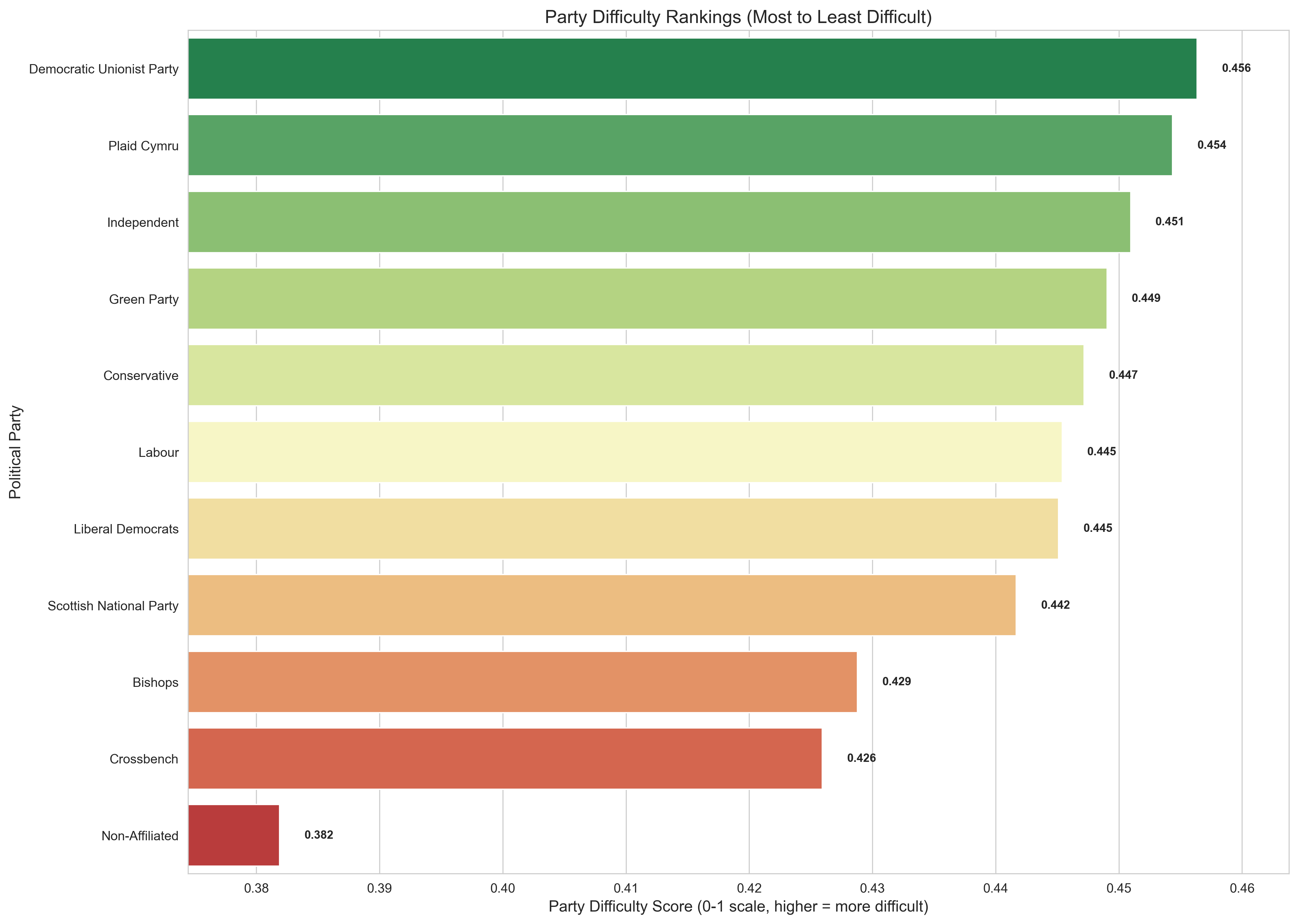}
  \caption{Party alignment difficulty scores (0-1 scale) for finetuned models across UK parliamentary parties. Dark green = most difficult}
  %\caption{Party alignment difficulty scores (0-1 scale). Darker = more difficult.}
  \label{fig:party_difficulty}
\end{figure}

\textbf{Party alignment Difficulty Analysis}. Applying cross-context stability analysis (Eq.\ref{eq:stability}), party difficulty scores ranged narrowly (0.382-0.456), with no statistically significant differences. This suggests relatively consistent modeling challenges across parties regardless of size or ideological composition. Results are presented in Figure \ref{fig:party_difficulty}. Detailed scores in Table~\ref{tab:party_difficulty_rankings}, Appendix~\ref{app:statistical_tests}.

\subsection{Topic Results}
\begin{figure}[ht]
  \centering
  \includegraphics[width=0.9\columnwidth,keepaspectratio]{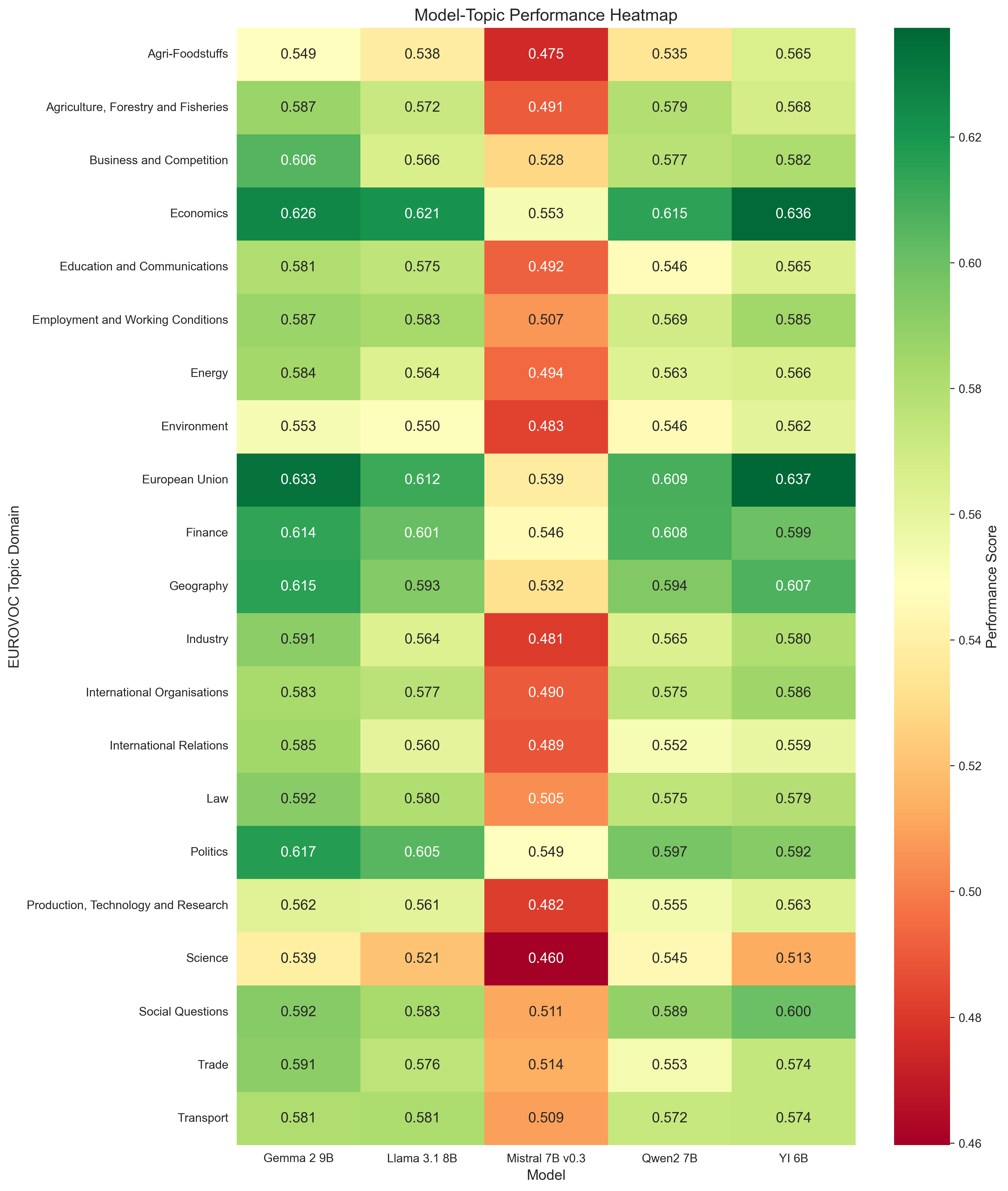}
  \caption{Model performance scores (0-1 scale) for finetuned models across topic domains. Color intensity indicates performance level}% (green = high, red = low).}
  %\caption{Model performance scores (0-1 scale) across topic domains.}
  \label{fig:topic_performance}
\end{figure}

\textbf{Topic Performance Patterns}. Figure \ref{fig:topic_performance} shows model performance across topic domains. Science achieved lowest scores (avg 0.516), while Economics (0.610) and European Union (0.606) showed highest performance. %
Detailed scores in Table~\ref{tab:topic_performance}, Appendix~\ref{app:statistical_tests}.
%(Detailed scores in the supplementary material.)

\begin{figure}[ht]
  \centering
  \includegraphics[width=0.9\columnwidth,keepaspectratio]{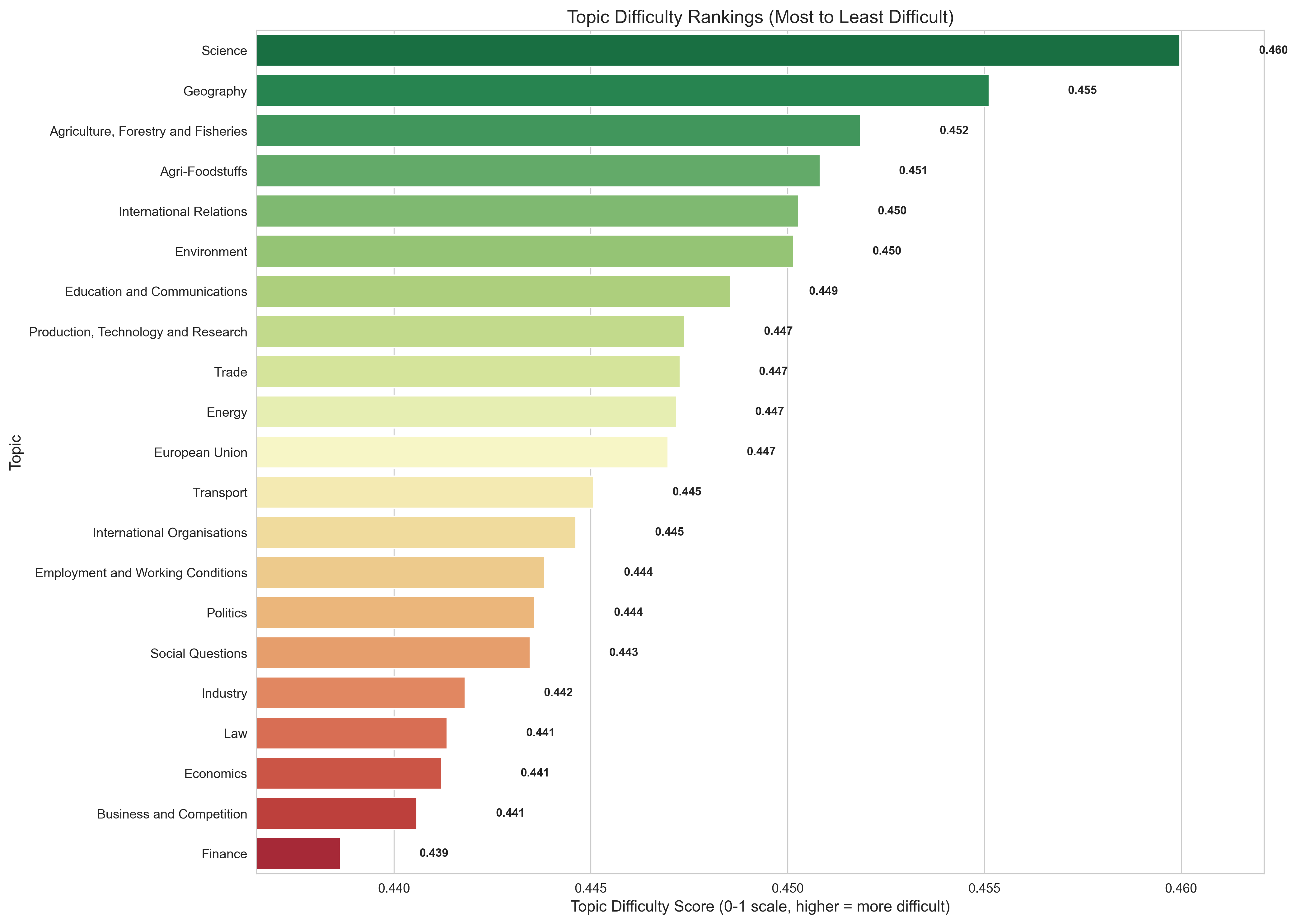}
  \caption{Topic difficulty scores (0-1 scale) for finetuned models across topic domains.}% Dark green = most difficult}
 %\caption{Topic difficulty scores (0-1 scale). Darker = more difficult.}
  \label{fig:topic_difficulty}
\end{figure}

\textbf{Topic Difficulty Analysis}. Different topics posed different challenges (Figure \ref{fig:topic_difficulty}). Science and Geography ranked as most difficult while Finance, Business, and Economics ranked lowest. Technical and natural science domains display higher cross-model disagreement than economic and political topics, consistent with greater terminological specialization and rapidly evolving concepts. In contrast, economic and political discussions employs more stable conceptual frameworks aligned with core parliamentary functions.

\subsection{Political Orientations Results}
\begin{figure}[ht]
  \centering
  \includegraphics[width=0.9\columnwidth,keepaspectratio]{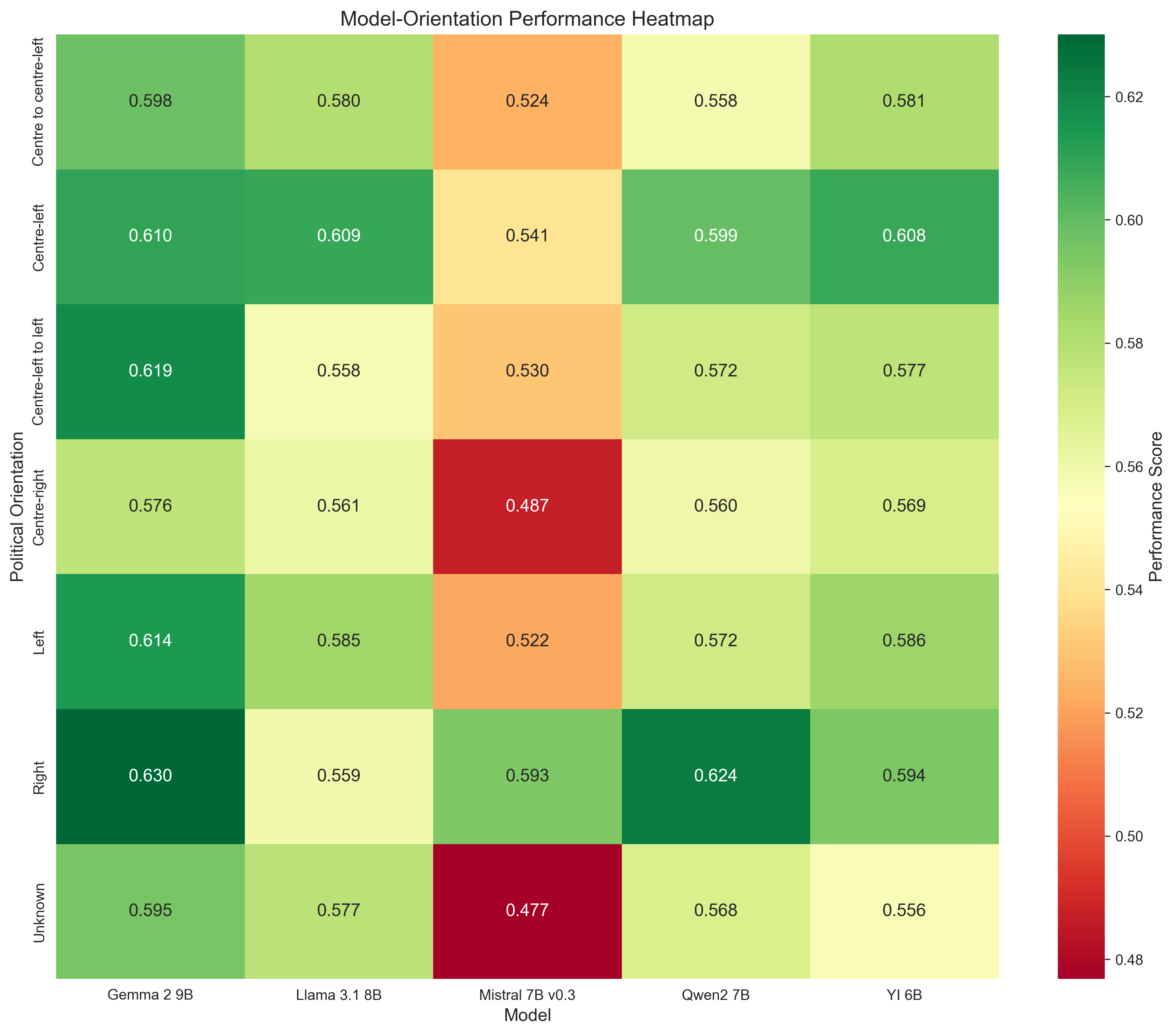}
  \caption{Model performance scores (0-1 scale) for finetuned models across political orientations. Color intensity indicates performance level}% (green = high, red = low).}
  %\caption{Model performance scores (0-1 scale) across political orientations.}
  \label{fig:orientation_performance}
\end{figure}

Performance across political orientations showed expected patterns. Centrist positions (Centre-left: 0.607, Centre-right: 0.551) dominated the dataset (88\%) and achieved higher scores. Model-specific strengths emerged as both \textit{Gemma} and \textit{Qwen}  achieved highest scores on Right positions and \textit{Mistral} underperformed consistently, indicating architectural rather than ideological limitations. As models are optimized for mainstream parliamentary speeches, extreme positions may require specialized training approaches. Figure \ref{fig:orientation_performance} illustrates these patterns. Detailed orientation difficulty rankings are provided in Table~\ref{tab:orientation_difficulty}, Appendix~\ref{app:statistical_tests}

 \subsection{Discussion}
Results establish several key findings: (i) Architectural design impacts political authenticity, with extended context windows enabling consistent improvements; (ii) Domain-specific fine-tuning proves essential as 45 of 70 metric comparisons showed statistically significant improvements and (iii) Novel political authenticity metrics (PSA, Party Align) successfully capture dimensions unavailable to conventional NLP metrics, validated through both fine-tuning responsiveness and discrimination testing (both $p<0.001$).

%% file: tables/performance_table_with_significance2.tex
\begin{table*}[t!]
    \centering
    %\small
    \resizebox{\textwidth}{!}{ % Ensure it fits the page width
    \begin{tabular}{l *{14}{c}}
        \toprule
        \textbf{Model} & \multicolumn{5}{c}{\textbf{Linguistic Quality}} & \multicolumn{4}{c}{\textbf{Semantic Coherence}} & \multicolumn{5}{c}{\textbf{Political Authenticity}} \\
        \cmidrule(lr){2-6} \cmidrule(lr){7-10} \cmidrule(lr){11-15}
        % This row contains the actual metric names and directions
        & \textbf{PPL} $\downarrow$ & \textbf{Dist-N} $\uparrow$ & \textbf{Self-BLEU} $\downarrow$ & \textbf{J\textunderscore Coh} $\uparrow$ & \textbf{J\textunderscore Conc} $\uparrow$
        & \textbf{GRUEN} $\uparrow$ & \textbf{BERTScore} $\uparrow$ & \textbf{MoverScore} $\uparrow$ & \textbf{J\textunderscore Rel} $\uparrow$
        & \textbf{PSA} $\uparrow$ & \textbf{Party Align} $\uparrow$ & \textbf{J\textunderscore Auth} $\uparrow$ & \textbf{J\textunderscore PolApp} $\uparrow$ & \textbf{J\textunderscore Qual} $\uparrow$ \\
        \midrule
Llama 3.1 8B (B) & 61.150 {\tiny $\pm 35.890$} & 0.988 {\tiny $\pm 0.019$} & 0.006 {\tiny $\pm 0.003$} & 0.591 {\tiny $\pm 0.085$} & 0.799 {\tiny $\pm 0.014$} & 0.504 {\tiny $\pm 0.007$} & 0.397 {\tiny $\pm 0.141$} & 0.502 {\tiny $\pm 0.134$} & 7.043 {\tiny $\pm 2.481$} & 5.889 {\tiny $\pm 3.101$} & 5.518 {\tiny $\pm 2.948$} & 4.364 {\tiny $\pm 2.647$} & 6.184 {\tiny $\pm 2.585$} & 4.796 {\tiny $\pm 2.403$} \\
Llama 3.1 8B (F) & 31.623$^{*}$ {\tiny $\pm 8.251$} & 0.974 {\tiny $\pm 0.018$} & 0.018 {\tiny $\pm 0.011$} & 0.537 {\tiny $\pm 0.146$} & 0.814$^{*}$ {\tiny $\pm 0.010$} & 0.509$^{*}$ {\tiny $\pm 0.006$} & 0.488$^{*}$ {\tiny $\pm 0.139$} & 0.576$^{*}$ {\tiny $\pm 0.107$} & 7.902$^{*}$ {\tiny $\pm 1.142$} & 7.125$^{*}$ {\tiny $\pm 2.608$} & 6.186$^{*}$ {\tiny $\pm 2.543$} & 6.133$^{*}$ {\tiny $\pm 3.001$} & 7.277$^{*}$ {\tiny $\pm 1.577$} & 5.415$^{*}$ {\tiny $\pm 2.235$} \\
\midrule
Gemma 2 9B (B) & 89.783 {\tiny $\pm 47.619$} & 0.992 {\tiny $\pm 0.007$} & 0.008 {\tiny $\pm 0.004$} & 0.579 {\tiny $\pm 0.074$} & 0.800 {\tiny $\pm 0.016$} & 0.508 {\tiny $\pm 0.007$} & 0.443 {\tiny $\pm 0.142$} & 0.542 {\tiny $\pm 0.118$} & 7.782 {\tiny $\pm 1.312$} & 4.756 {\tiny $\pm 3.361$} & 5.822 {\tiny $\pm 2.473$} & 3.842 {\tiny $\pm 2.357$} & 6.513 {\tiny $\pm 2.044$} & 4.469 {\tiny $\pm 1.987$} \\
Gemma 2 9B (F) & 102.382 {\tiny $\pm 60.592$} & 0.990 {\tiny $\pm 0.021$} & 0.010 {\tiny $\pm 0.006$} & 0.532 {\tiny $\pm 0.078$} & 0.799 {\tiny $\pm 0.018$} & 0.508 {\tiny $\pm 0.006$} & 0.497$^{*}$ {\tiny $\pm 0.137$} & 0.589$^{*}$ {\tiny $\pm 0.102$} & 7.502 {\tiny $\pm 1.487$} & 4.970 {\tiny $\pm 3.489$} & 5.598 {\tiny $\pm 2.463$} & 4.215$^{*}$ {\tiny $\pm 2.529$} & 7.314$^{*}$ {\tiny $\pm 1.617$} & 4.976$^{*}$ {\tiny $\pm 1.915$} \\
\midrule
Mistral 7B v0.3 (B) & 31.428 {\tiny $\pm 18.844$} & 0.966 {\tiny $\pm 0.052$} & 0.008 {\tiny $\pm 0.005$} & 0.598 {\tiny $\pm 0.108$} & 0.806 {\tiny $\pm 0.012$} & 0.505 {\tiny $\pm 0.007$} & 0.417 {\tiny $\pm 0.146$} & 0.522 {\tiny $\pm 0.131$} & 6.582 {\tiny $\pm 2.963$} & 6.850 {\tiny $\pm 3.212$} & 5.428 {\tiny $\pm 3.184$} & 4.234 {\tiny $\pm 2.780$} & 5.627 {\tiny $\pm 2.732$} & 4.182 {\tiny $\pm 2.647$} \\
Mistral 7B v0.3 (F) & 29.379$^{*}$ {\tiny $\pm 10.852$} & 0.972$^{*}$ {\tiny $\pm 0.019$} & 0.017 {\tiny $\pm 0.012$} & 0.590 {\tiny $\pm 0.082$} & 0.820$^{*}$ {\tiny $\pm 0.009$} & 0.506$^{*}$ {\tiny $\pm 0.006$} & 0.437$^{*}$ {\tiny $\pm 0.139$} & 0.506 {\tiny $\pm 0.122$} & 7.954$^{*}$ {\tiny $\pm 1.085$} & 8.942$^{*}$ {\tiny $\pm 2.051$} & 5.744$^{*}$ {\tiny $\pm 2.904$} & 3.986 {\tiny $\pm 2.735$} & 6.388$^{*}$ {\tiny $\pm 2.097$} & 3.745 {\tiny $\pm 2.089$} \\
\midrule
Qwen2 7B (B) & 44.927 {\tiny $\pm 21.739$} & 0.981 {\tiny $\pm 0.026$} & 0.020 {\tiny $\pm 0.014$} & 0.534 {\tiny $\pm 0.142$} & 0.799 {\tiny $\pm 0.009$} & 0.507 {\tiny $\pm 0.006$} & 0.444 {\tiny $\pm 0.140$} & 0.562 {\tiny $\pm 0.114$} & 7.919 {\tiny $\pm 1.222$} & 5.937 {\tiny $\pm 1.652$} & 6.938 {\tiny $\pm 2.345$} & 6.558 {\tiny $\pm 2.583$} & 7.324 {\tiny $\pm 1.751$} & 6.369 {\tiny $\pm 1.627$} \\
Qwen2 7B (F) & 36.040$^{*}$ {\tiny $\pm 9.144$} & 0.982 {\tiny $\pm 0.013$} & 0.017$^{*}$ {\tiny $\pm 0.009$} & 0.570$^{*}$ {\tiny $\pm 0.086$} & 0.815$^{*}$ {\tiny $\pm 0.009$} & 0.510$^{*}$ {\tiny $\pm 0.006$} & 0.489$^{*}$ {\tiny $\pm 0.135$} & 0.573$^{*}$ {\tiny $\pm 0.106$} & 8.056$^{*}$ {\tiny $\pm 0.887$} & 7.617$^{*}$ {\tiny $\pm 2.697$} & 6.081 {\tiny $\pm 2.641$} & 5.718 {\tiny $\pm 3.017$} & 7.146 {\tiny $\pm 1.617$} & 5.015 {\tiny $\pm 2.100$} \\
\midrule
YI 6B (B) & 81.424 {\tiny $\pm 56.676$} & 0.990 {\tiny $\pm 0.013$} & 0.006 {\tiny $\pm 0.005$} & 0.616 {\tiny $\pm 0.052$} & 0.794 {\tiny $\pm 0.019$} & 0.503 {\tiny $\pm 0.007$} & 0.344 {\tiny $\pm 0.144$} & 0.424 {\tiny $\pm 0.150$} & 6.737 {\tiny $\pm 2.608$} & 4.322 {\tiny $\pm 3.652$} & 4.503 {\tiny $\pm 2.864$} & 2.969 {\tiny $\pm 1.965$} & 5.387 {\tiny $\pm 2.634$} & 3.100 {\tiny $\pm 2.118$} \\
YI 6B (F) & 42.835$^{*}$ {\tiny $\pm 13.527$} & 0.986 {\tiny $\pm 0.011$} & 0.016 {\tiny $\pm 0.009$} & 0.566 {\tiny $\pm 0.082$} & 0.812$^{*}$ {\tiny $\pm 0.010$} & 0.509$^{*}$ {\tiny $\pm 0.006$} & 0.492$^{*}$ {\tiny $\pm 0.139$} & 0.581$^{*}$ {\tiny $\pm 0.102$} & 8.039$^{*}$ {\tiny $\pm 0.746$} & 6.838$^{*}$ {\tiny $\pm 2.822$} & 6.062$^{*}$ {\tiny $\pm 2.555$} & 6.128$^{*}$ {\tiny $\pm 2.867$} & 7.334$^{*}$ {\tiny $\pm 1.479$} & 5.409$^{*}$ {\tiny $\pm 2.061$} \\
\bottomrule
\end{tabular}
    }
    \caption{Complete Performance Comparison between Baseline (B) and Fine-Tuned (F) Models across Evaluation Metrics. Asterisks (*) indicate statistically significant improvements (p < 0.05).}
    \label{tab:performance_table} 
\end{table*} 

%% file: 7._conclusion.tex
\section{Discussion and Conclusion}

To the best of our knowledge, ParliaBench represents the first benchmark resource addressing the specialized challenges of parliamentary speech generation, comprising dataset, evaluation framework, novel metrics, and baseline benchmarks. ParliaBench provides standardized evaluation protocols and baseline performance results that 
support systematic comparison and reproducible research in the field.
Our results demonstrate that domain specific fine-tuning produces significant quality improvements, while 
our novel political authenticity metrics successfully capture ideological dimensions absent from conventional evaluation approaches.

\textbf{Future Directions:} Extensions include: (i) multilingual evaluation for European parliamentary systems, (ii) human evaluation protocols for validation, and (iii) systematic assessment of political bias and perspective maintenance across viewpoints.

%% file: appendix_a.tex
% !TEX root = paper.tex
\section{Dataset Details} \label{appendix:dataset}

\subsection{Dataset Statistics}

Table \ref{tab:party_dist} presents detailed breakdown of speeches by political affiliation in the Parliamentary Debates Benchmark.

\begin{table}[h]
\centering
\begin{tabular}{llrrr}
\toprule
\textbf{Party} & \textbf{Orientation} & \textbf{Speeches} & \textbf{Speakers} & \textbf{Active Period} \\
\midrule
Conservative & Centre-right & 263,513 & 792 & 2015--2022 \\
Labour & Centre-left & 108,831 & 592 & 2015--2022 \\
Scottish National Party & Centre-left & 23,562 & 67 & 2015--2022 \\
Liberal Democrats & Centre to centre-left & 23,517 & 168 & 2015--2022 \\
Crossbench & Unknown & 11,878 & 215 & 2015--2022 \\
Democratic Unionist Party & Right & 6,610 & 15 & 2015--2022 \\
Independent & Unknown & 2,783 & 45 & 2015--2022 \\
Plaid Cymru & Centre-left to left & 2,229 & 7 & 2015--2022 \\
Green Party & Left & 1,992 & 3 & 2015--2022 \\
Non-Affiliated & Unknown & 1,713 & 60 & 2015--2022 \\
Bishops & Unknown & 1,150 & 41 & 2015--2022 \\
\midrule
\textbf{Total} & & \textbf{447,778} & \textbf{1901} & \\
\bottomrule
\end{tabular}
\caption{Political Parties Represented in Dataset}
\label{tab:party_dist}
\end{table}

\subsection{Hybrid Classification Strategy} \label{appendix:mapping}

This subsection documents the hybrid classification strategy employed to map Comparative Agendas Project (CAP) categories to EuroVoc domains for topic assignment in our parliamentary speech dataset. We employed direct semantic mapping for 16 categories, while 6 remaining categories required automated classification, as detailed in Table \ref{tab:cap_eurovoc_mapping}.

\begin{table}[h]
\centering
\begin{tabular}{llll}
\hline
\textbf{CAP Category} & \textbf{EuroVoc Category} & \textbf{Method} & \textbf{Rationale} \\
\hline
Energy & Energy & Direct & Exact match \\
Environment & Environment & Direct & Exact match \\
Transportation & Transport & Direct & Exact match \\
Foreign Trade & Trade & Direct & Clear semantic alignment \\
Government Operations & Politics & Direct & Clear semantic alignment \\
International Affairs & International Relations & Direct & Clear semantic alignment \\
Labor & Employment and Working Conditions & Direct & Clear semantic alignment \\
Law and Crime & Law & Direct & Clear semantic alignment \\
Agriculture & Agriculture, Forestry, Fisheries & Direct & Clear semantic alignment \\
Education & Education and Communications & Direct & Clear semantic alignment \\
Macroeconomics & Economy & Direct & Clear semantic alignment \\
Social Welfare & Social Questions & Direct & Clear semantic alignment \\
Technology & Production, Technology and Research & Direct & Clear semantic alignment \\
Civil Rights & --- & Automated & Complex assignment \\
Domestic Commerce & --- & Automated & Complex assignment \\
Culture & --- & Automated & Complex assignment \\
Health & --- & Automated & Complex assignment \\
Defense & --- & Automated & Complex assignment \\
Housing & --- & Automated & Complex assignment \\
Immigration & --- & Automated & Complex assignment \\
Public Lands & --- & Automated & Complex assignment \\
Mix/Other & --- & Automated & Complex assignment \\
\hline
\end{tabular}
\caption{Mapping between CAP Categories and EuroVoc Categories}
\label{tab:cap_eurovoc_mapping}
\end{table}

%% file: appendix_b.tex
\section{LLM-as-a-Judge Evaluation Prompts} \label{appendix:llm_judge_prompts}

This appendix documents the automated evaluation system used to assess the quality of generated parliamentary speeches. The system employs Flow-Judge-v0.1, a 3.8B parameter evaluation model, to score speeches across six dimensions using a 10-point scale.

\subsection{Evaluation Framework}

All speeches are evaluated on six metrics:
\begin{enumerate}[itemsep=2pt, topsep=5pt, parsep=0pt]
    \item \textbf{Coherence (j\_coh)}: Logical flow and structural clarity
    \item \textbf{Conciseness (j\_conc)}: Efficiency of expression within parliamentary context
    \item \textbf{Relevance (j\_rel)}: Coverage of prompt requirements
    \item \textbf{Authenticity (j\_auth)}: Naturalness of Westminster discourse
    \item \textbf{Political Appropriateness (j\_polapp)}: Alignment with party positions
    \item \textbf{Overall Quality (j\_qual)}: Persuasiveness and argumentation strength
\end{enumerate}

Each metric uses a structured prompt template with explicit evaluation criteria, a 10-point scoring rubric, and detailed instructions for the judge model. The system processes speeches in batches of 32 for computational efficiency.

\subsection{Metric 1: Coherence}
\label{app:metric-coherence}

\subsubsection{Evaluation Criteria}
Does the speech follow a clear logical progression? Are arguments well-connected and ideas flow naturally with appropriate parliamentary structure?

\subsubsection{Prompt Template and Scoring Rubric}
\begin{tcolorbox}[templatebox, breakable, title={Coherence Evaluation Prompt}]
\begin{verbatim}
# GOAL
You are a senior parliamentary discourse analyst with 15 years of 
experience evaluating political speech quality in Westminster-style 
parliaments. Your task is to evaluate the coherence (logical flow 
and structure) of this UK parliamentary speech.

# OUTPUT
Below is the parliamentary speech to evaluate:
<output>
{{ANSWER}}
</output>

# EVALUATION CRITERIA AND SCORING RUBRIC
<evaluation_criteria>
Does the speech follow a clear logical progression? Are arguments 
well-connected? Is parliamentary structure maintained?
</evaluation_criteria>

<scoring_rubric>
- Score 1-2: Incoherent. No discernible logical structure, 
  fragmented or nonsensical.
- Score 3-4: Poor coherence. Disjointed arguments, unclear 
  progression, difficult to follow main point.
- Score 5-6: Adequate coherence. Main argument identifiable but 
  weak connections, some repetition or tangents.
- Score 7-8: Strong coherence. Mostly logical progression with 
  minor transition issues, follows parliamentary structure.
- Score 9-10: Exceptional logical flow. Each point builds naturally, 
  clear thesis and conclusion, perfect parliamentary structure.
</scoring_rubric>

# INSTRUCTIONS FOR THE EVALUATION
1. Before scoring, analyze the speech structure step-by-step.
2. Quote specific evidence or patterns from the speech supporting 
   your assessment.
3. Use the full 1-10 range - don't default to middle scores (5-7) 
   without justification.
4. Pay attention to how well arguments connect and whether 
   parliamentary conventions are maintained.

## FORMAT FOR THE EVALUATION
- Write your verbal feedback inside <feedback> tags without any 
  surrounding text.
- Write the numeric score inside <score> tags, always after the 
  feedback.

Please accurately evaluate the speech. Strictly adhere to the 
evaluation criteria and rubric.
\end{verbatim}
\end{tcolorbox}

\subsection{Metric 2: Conciseness}
\label{app:metric-conciseness}

\subsubsection{Evaluation Criteria}
Is the message conveyed efficiently without excessive verbosity, within the context of formal parliamentary speech where longer discourse is expected?

\subsubsection{Prompt Template and Scoring Rubric}
\begin{tcolorbox}[templatebox, breakable, title={Conciseness Evaluation Prompt}]
\begin{verbatim}
# GOAL
You are a senior parliamentary discourse analyst with 15 years of 
experience evaluating political speech quality. Your task is to 
evaluate the conciseness of this UK parliamentary speech using 
parliamentary standards (where formal speeches are naturally longer 
than everyday communication).

# OUTPUT
Below is the parliamentary speech to evaluate:
<output>
{{ANSWER}}
</output>

# EVALUATION CRITERIA AND SCORING RUBRIC
<evaluation_criteria>
Does the speech convey its message efficiently without unnecessary 
verbosity? Remember: Parliamentary speeches are naturally more 
formal and verbose - assess within this context.
</evaluation_criteria>

<scoring_rubric>
- Score 1-2: Extremely verbose. Main argument lost among unrelated 
  statements, excessive repetition, rambling.
- Score 3-4: Verbose. Significant digression into unrelated matters, 
  could be 30-50% shorter while preserving intent.
- Score 5-6: Adequate. Main point clear but diluted by length or 
  tangential references, some ceremonial language adds weight.
- Score 7-8: Concise. Clear focus with minor digressions 
  (constituency examples), formal niceties don't impede core message.
- Score 9-10: Highly concise for parliamentary standard. Focused 
  and disciplined, each point supports main argument, courtesy 
  phrases used sparingly.
</scoring_rubric>

# INSTRUCTIONS FOR THE EVALUATION
1. Identify the main argument and check for focus vs digressions.
2. Assess if courtesy and formality serve rhetorical purpose 
   (parliamentary context).
3. Detect unnecessary repetition or verbose phrasing.
4. Remember: Parliamentary speeches are naturally longer than 
   everyday communication - adjust expectations accordingly.
5. Quote examples showing appropriate length or excessive verbosity.

## FORMAT FOR THE EVALUATION
- Write your verbal feedback inside <feedback> tags without any 
  surrounding text.
- Write the numeric score inside <score> tags, always after the 
  feedback.

Please accurately evaluate the speech. Strictly adhere to the 
evaluation criteria and rubric.
\end{verbatim}
\end{tcolorbox}

\subsection{Metric 3: Relevance}
\label{app:metric-relevance}

\subsubsection{Evaluation Criteria}
Does the speech directly address the prompt/question and cover all core concerns specified in the instruction?

\subsubsection{Prompt Template and Scoring Rubric}

\begin{tcolorbox}[templatebox, breakable, title={Relevance Evaluation Prompt}]
\begin{verbatim}
# GOAL
You are a parliamentary procedures expert with 15 years analyzing 
parliamentary questions and responses. Your task is to evaluate the 
relevance of this UK parliamentary speech - whether it directly 
addresses the prompt/question asked and covers core concerns.

# OUTPUT
Below is the parliamentary speech to evaluate:
<output>
{{ANSWER}}
</output>

# EVALUATION CRITERIA AND SCORING RUBRIC
<evaluation_criteria>
Does the speech directly address the prompt/question asked and cover 
the core concerns? Prompt was: "{{PROMPT}}"
</evaluation_criteria>

<scoring_rubric>
- Score 1-2: Non-answer. Completely ignores prompt, discusses 
  unrelated topics.
- Score 3-4: Weak answer. Barely addresses prompt (<50%), mostly 
  generic, significant irrelevant content.
- Score 5-6: Partial answer. Addresses main topic but misses key 
  aspects (50-75%), some tangents.
- Score 7-8: Strong answer. Addresses MOST aspects (75%+), provides 
  specifics, minimal irrelevant content.
- Score 9-10: Comprehensive answer. Addresses ALL aspects with 
  specific details, no tangents, complete coverage.
</scoring_rubric>

# INSTRUCTIONS FOR THE EVALUATION
1. Identify what specifically is being asked in the prompt.
2. Check coverage of each requirement mentioned in the prompt.
3. Assess specificity (concrete details vs vague generalities).
4. Detect tangents and estimate percentage of off-topic content.
5. For generic prompts, evaluate if response matches expected 
   political context.
6. Quote evidence showing relevance or lack thereof.

## FORMAT FOR THE EVALUATION
- Write your verbal feedback inside <feedback> tags without any 
  surrounding text.
- Write the numeric score inside <score> tags, always after the 
  feedback.

Please accurately evaluate the speech. Strictly adhere to the 
evaluation criteria and rubric.
\end{verbatim}
\end{tcolorbox}

\subsection{Metric 4: Authenticity}
\label{app:metric-authenticity}

\subsubsection{Evaluation Criteria}
Does this sound like genuine Westminster discourse or AI-generated text? Consider language patterns, rhetorical devices, and parliamentary conventions.

\subsubsection{Prompt Template and Scoring Rubric}
\begin{tcolorbox}[templatebox, breakable, title={Authenticity Evaluation Prompt}]
\begin{verbatim}
# GOAL
You are a political discourse specialist with 20 years of experience 
studying UK parliamentary speeches, with deep knowledge of party 
ideologies and rhetorical patterns. Your task is to evaluate whether 
this speech sounds like genuine Westminster discourse or AI-generated 
text.

# OUTPUT
Below is the parliamentary speech to evaluate:
<output>
{{ANSWER}}
</output>

# EVALUATION CRITERIA AND SCORING RUBRIC
<evaluation_criteria>
Does this sound like real parliamentary speech or AI-generated text? 
Are parliamentary conventions natural or forced? Is the language 
authentic to Westminster discourse? Context: Party={{PARTY}}, 
Topic={{TOPIC}}
</evaluation_criteria>

<scoring_rubric>
- Score 1-2: Completely artificial. Marketing-like or chatbot-like 
  with obvious AI patterns, no parliamentary structure.
- Score 3-4: Obviously artificial. Generic corporate-speak, 
  essay-like writing, missing or forced parliamentary conventions.
- Score 5-6: Acceptable but noticeably artificial. Template-like 
  but correct format, sounds scripted with stiff language.
- Score 7-8: Convincing. Mostly natural parliamentary discourse 
  with minor artificiality, authentic flow overall.
- Score 9-10: Highly authentic. Indistinguishable from real speech, 
  natural conventions, party-appropriate rhetoric.
</scoring_rubric>

# INSTRUCTIONS FOR THE EVALUATION
1. Read the speech twice: first for overall impression, second for 
   detailed analysis.
2. Check if "Mr. Speaker" and appropriate House references are used 
   naturally (not forced).
3. Assess if language sounds natural or scripted/polished.
4. Detect AI patterns: generic phrases, perfect grammar, 
   corporate-speak, marketing tone.
5. Quote specific evidence showing authenticity or artificiality.

## FORMAT FOR THE EVALUATION
- Write your verbal feedback inside <feedback> tags without any 
  surrounding text.
- Write the numeric score inside <score> tags, always after the 
  feedback.

Please accurately evaluate the speech. Strictly adhere to the 
evaluation criteria and rubric.
\end{verbatim}
\end{tcolorbox}
%\end{verbatim}

\subsection{Metric 5: Political Appropriateness}
\label{app:metric-political}

\subsubsection{Evaluation Criteria}
Does the stance align with the specified party's typical positions and rhetoric on this topic?

\subsubsection{Prompt Template and Scoring Rubric}
%\begin{verbatim}
\begin{tcolorbox}[templatebox, breakable, title={Political Appropriateness Evaluation Prompt}]
\begin{verbatim}
# GOAL
You are a political discourse specialist with 20 years studying UK 
parliamentary speeches and party ideologies. Your task is to evaluate 
political appropriateness - whether the speech's stance aligns with 
the party's typical positions and uses appropriate Westminster 
parliamentary conventions.

# OUTPUT
Below is the parliamentary speech to evaluate:
<output>
{{ANSWER}}
</output>

# EVALUATION CRITERIA AND SCORING RUBRIC
<evaluation_criteria>
Does the stance align with the party's typical positions on this 
topic? Does the language and framing match Westminster parliamentary 
conventions? Context: Party={{PARTY}}, Topic={{TOPIC}}
</evaluation_criteria>

<scoring_rubric>
- Score 1-2: Completely misaligned. Opposes party's core positions 
  or uses rival party's rhetoric.
- Score 3-4: Poor alignment. Key positions contradict party stance 
  or inappropriate framing for the party.
- Score 5-6: Somewhat aligned. Generic positions that are slightly 
  off-brand or could be any party.
- Score 7-8: Good alignment. Mostly consistent with party stance, 
  language matches party's typical rhetoric.
- Score 9-10: Perfect alignment. Language and framing exactly match 
  party's typical positions and rhetoric.
</scoring_rubric>

# INSTRUCTIONS FOR THE EVALUATION
1. Consider the party's known positions on this topic based on their 
   political orientation.
2. Check if speech aligns with expected party stance and framing.
3. Assess if language and rhetoric are appropriate for this specific 
   party.
4. Detect contradictions or use of opposing party's rhetoric.
5. Quote evidence showing appropriate or inappropriate party 
   alignment.

## FORMAT FOR THE EVALUATION
- Write your verbal feedback inside <feedback> tags without any 
  surrounding text.
- Write the numeric score inside <score> tags, always after the 
  feedback.

Please accurately evaluate the speech. Strictly adhere to the 
evaluation criteria and rubric.
\end{verbatim}
\end{tcolorbox}

\subsection{Metric 6: Overall Quality}
\label{app:metric-quality}

\subsubsection{Evaluation Criteria}
How effective is this as political communication? Is the argumentation sophisticated and persuasive?

\subsubsection{Prompt Template and Scoring Rubric}
%\begin{verbatim}
\begin{tcolorbox}[templatebox, breakable, title={Overall Quality Evaluation Prompt}]
\begin{verbatim}
# GOAL
You are a political discourse specialist with 20 years studying UK 
parliamentary speeches. Your task is to evaluate the overall quality 
of this UK parliamentary speech - its effectiveness as political 
communication, persuasiveness, and sophistication.

# OUTPUT
Below is the parliamentary speech to evaluate:
<output>
{{ANSWER}}
</output>

# EVALUATION CRITERIA AND SCORING RUBRIC
<evaluation_criteria>
How effective is this speech as political communication? Would it 
genuinely influence parliamentary debate? Is the argumentation 
sophisticated and compelling? Context: Party={{PARTY}}, 
Topic={{TOPIC}}
</evaluation_criteria>

<scoring_rubric>
- Score 1-2: Very poor. Nonsensical arguments, no persuasive value, 
  would damage speaker's credibility.
- Score 3-4: Weak. Poor argumentation, logic gaps, lacks evidence, 
  unconvincing presentation.
- Score 5-6: Adequate. Makes its point but not compelling, basic 
  arguments without depth, functional but forgettable.
- Score 7-8: Strong. Clear arguments with good evidence, 
  professional and convincing, respectable contribution.
- Score 9-10: Exceptional. Sophisticated argumentation, compelling 
  evidence, memorable phrasing, would genuinely influence debate.
</scoring_rubric>

# INSTRUCTIONS FOR THE EVALUATION
1. Assess argumentation quality: Is logic sound? Are claims 
   supported?
2. Evaluate evidence: Does speech use specific data, examples, or 
   reasoning?
3. Check rhetorical devices: Are there memorable phrases or 
   effective persuasion techniques?
4. Consider potential impact: Would this speech influence other 
   parliamentarians?
5. Be tough: Real parliamentary speeches often score 6-8, not 9-10.
6. Quote evidence showing strong or weak argumentation.

## FORMAT FOR THE EVALUATION
- Write your verbal feedback inside <feedback> tags without any 
  surrounding text.
- Write the numeric score inside <score> tags, always after the 
  feedback.

Please accurately evaluate the speech. Strictly adhere to the 
evaluation criteria and rubric.
\end{verbatim}
\end{tcolorbox}

\subsection{Implementation Details}
\label{app:implementation}

\subsubsection{Model Configuration}
The evaluation system uses the following configuration:
\begin{itemize}[itemsep=2pt, topsep=5pt, parsep=0pt]
    \item \textbf{Judge Model}: Flow-Judge-v0.1 (3.8B parameters, 4-bit quantization)
    \item \textbf{Batch Size}: 32 speeches per batch
    \item \textbf{Temperature}: 0.3 (for consistent scoring)
    \item \textbf{Max New Tokens}: 2000 (for detailed feedback)
    \item \textbf{Default Score}: -1 (for missing/error values)
\end{itemize}

%% file: appendix_c.tex
\section{Setup Implementation Details} \label{appendix:setup}

\subsection{Model Architecture Details}
\begin{table}[!htbp]
\centering
\small
\setlength{\tabcolsep}{3pt}
\begin{tabular}{lccccc}
\toprule
\textbf{Model} & \textbf{Quantized Version} & \textbf{Memory Reduction} & \textbf{Inference Speed} & \textbf{Key Features} \\
\midrule
Mistral 7B v0.3 & mistral-7b-v0.3-bnb-4bit & 62\% & 2.2× & GQA, SWA \\
Llama 3.1 8B & Meta-Llama-3.1-8B-bnb-4bit & 58\% & 2.4× & 128k context \\
Gemma 2 9B & gemma-2-9b-bnb-4bit & 58\% & 2.2× & Alternating attention \\
Qwen2 7B & Qwen2-7B-bnb-4bit & N/A & N/A & Multilingual \\
Falcon-H1 7B & falcon-7b-bnb-4bit & N/A & N/A & Training efficiency \\
\bottomrule
\end{tabular}
\caption{Model Technical Specifications and Quantization Details}
\label{tab:model_specs}
\end{table}
Table \ref{tab:model_specs} summarizes technical specifications and quantization details for the selected models.

\subsection{QLoRA Configuration Details}

This section provides complete technical specifications for the Quantized Low-Rank Adaptation (QLoRA) implementation used across all model architectures.

\begin{table}[h]
\centering
\small % Reduce font size to fit
\begin{tabular}{p{0.3\columnwidth} p{0.25\columnwidth} p{0.35\columnwidth}}
\toprule
\textbf{Parameter} & \textbf{Value} & \textbf{Rationale} \\
\midrule
LoRA Rank (r) & 16 & Optimal balance for fast fine-tuning \\
LoRA Alpha & 16 & Set equal to rank ($\alpha/r=1$) for baseline \\
Target Modules & 7 layers & All linear transformations \\
LoRA Dropout & 0 & Enable Unsloth optimizations \\
Bias Configuration & none & Faster training, reduced memory \\
Random State & 3407 & Reproducibility across architectures \\
\bottomrule
\end{tabular}
\caption{Complete QLoRA Configuration Parameters}
\label{tab:qlora_detailed}
\end{table}

The QLoRA configuration parameters were selected based on established best practices for parameter-efficient fine-tuning in specialized domains. The rank value of 16 provides sufficient adaptation capacity while maintaining computational efficiency. Setting LoRA Alpha equal to the rank ensures reliable baseline performance, while disabling dropout enables Unsloth framework optimizations essential for efficient training on A100 hardware.

Target modules encompass all linear transformation layers (q\_proj, k\_proj, v\_proj, o\_proj, gate\_proj, up\_proj, down\_proj) to achieve performance comparable to full fine-tuning while requiring only a fraction of the computational resources. The consistent random state across all architectures ensures reproducible results essential for systematic model comparison.

% \subsection{Training Configuration}
\begin{table}[h]
\centering
\small % Reduce font size to fit

\begin{tabular}{p{0.3\columnwidth} p{0.25\columnwidth} p{0.35\columnwidth}}
\toprule
\textbf{Parameter} & \textbf{Value} & \textbf{Justification} \\
\midrule
Batch Size & 64 & GPU memory optimization \\
Gradient Accumulation & 1 & No gradient accumulation \\
Learning Rate & 2e-4 & Standard for LoRA fine-tuning \\
Max Steps & 11194 & 2 epochs (Prevents overfitting) \\
Warmup Steps & 336 & 3\% of max steps  \\
Optimizer & adamw  & Memory-efficient \\
Weight Decay & 0.01 & Prevents overfitting on political data \\
Max Sequence Length & 1024 & Accommodates speech lengths \\
Scheduler & linear & Linear learning rate schedule\\
\bottomrule
\end{tabular}
\caption{Complete Training Configuration for Parliamentary Speech Generation}
\label{tab:training_config_detailed}
\end{table}

\subsection{Speech Generator Prompts} 
This section provides the complete chat template specifications used for training and generation across all model architectures, ensuring reproducibility of experimental results.

\subsubsection{Multi-Model Training Chat Template Architecture}
All models are trained with the following template: \\
\vspace{0.5cm} % leaves a 1 cm blank space

\texttt{You are a seasoned UK parliamentary member.
Use proper British parliamentary language appropriate for the specified House.}\\
\texttt{The speech should reflect the political orientation and typical positions of the specified party on the given topic.}\\

\vspace{0.5cm} % leaves a 1 cm blank space

\subsubsection{Speech Generation System Prompt}

We generated the speeches for the finetuned and the baseline models using the following prompt for all models. 
\vspace{0.5cm} 

 \texttt{You are a seasoned UK parliamentary member. Generate a coherent speech of 
min\_words - max\_words words in standard English (no Unicode artifacts, no special characters).} \\
\texttt{
Use proper British parliamentary language appropriate for the specified House. }\\
\texttt{The speech should reflect the political orientation and typical positions of the specified party on the given topic. }\\

\vspace{0.5cm} 

Each model architecture wraps this shared structure in its own special formatting tokens and control sequences.

\begin{tcolorbox}[templatebox, title={Mistral Model Template}]
\begin{verbatim}
<s>[INST] {SYSTEM_PROMPT}
Context: {context}
Instruction: {instruction} [/INST] {response}</s>
\end{verbatim}
\end{tcolorbox}

\begin{tcolorbox}[templatebox, title={Llama 3.1 Model Template}]
\begin{verbatim}
<|begin_of_text|><|start_header_id|>system<|end_header_id|>
{SYSTEM_PROMPT}
<|eot_id|><|start_header_id|>user<|end_header_id|>
Context: {context}
Instruction: {instruction}
<|eot_id|><|start_header_id|>assistant<|end_header_id|>
{response}<|eot_id|>
\end{verbatim}
\end{tcolorbox}

\begin{tcolorbox}[templatebox, title={Gemma 2 Model Template}]
\begin{verbatim}
<bos><start_of_turn>user
{SYSTEM_PROMPT}
Context: {context}
Instruction: {instruction}<end_of_turn>
<start_of_turn>model
{response}<end_of_turn>
\end{verbatim}
\end{tcolorbox}

\begin{tcolorbox}[templatebox, title={Qwen2 Model Template (ChatML format)}]
\begin{verbatim}
<|im_start|>system
{SYSTEM_PROMPT}
<|im_end|>
<|im_start|>user
Context: {context}
Instruction: {instruction}<|im_end|>
<|im_start|>assistant
{response}<|im_end|>
\end{verbatim}
\end{tcolorbox}

\begin{tcolorbox}[templatebox, title={Yi-1.5 Model Template (ChatML format)}]
\begin{verbatim}
<|im_start|>system
{SYSTEM_PROMPT}<|im_end|>
<|im_start|>user
Context: {context}
Instruction: {instruction}<|im_end|>
<|im_start|>assistant
{response}<|im_end|>
\end{verbatim}
\end{tcolorbox}

% \textbf{Mistral Model Template:}
%     \begin{quote}
%     \texttt{<s>[INST] SYSTEM\_PROMPT\\
%     Context: \{context\}\\
%     Instruction: \{instruction\} [/INST]}
%     \end{quote}
% \textbf{Llama 3.1 Model Template:}
%     \begin{quote}
%     \texttt{<|begin\_of\_text|><|start\_header\_id|>system<|end\_header\_id|>\\
%     SYSTEM\_PROMPT<|eot\_id|><|start\_header\_id|>user<|end\_header\_id|>\\
%     Context: \{context\}\\
%     Instruction: \{instruction\}<|eot\_id|><|start\_header\_id|>assistant<|end\_header\_id|>}
%     \end{quote}

% \textbf{Gemma 2 Model Template:}
%     \begin{quote}
%     \texttt{<bos><start\_of\_turn>user\\
%     SYSTEM\_PROMPT\\
%     Context: \{context\}\\
%     Instruction: \{instruction\}<end\_of\_turn>\\
%     <start\_of\_turn>model}
%     \end{quote}

% \textbf{Qwen2 Model Template (ChatML format):}
%     \begin{quote}
%     \texttt{<|im\_start|>system\\
%     SYSTEM\_PROMPT<|im\_end|>\\
%     <|im\_start|>user\\
%     Context: \{context\}\\
%     Instruction: \{instruction\}<|im\_end|>\\
%     <|im\_start|>assistant}
%     \end{quote}

% \textbf{Yi-1.5 Model Template (ChatML format):}
%     \begin{quote}
%     \texttt{<|im\_start|>system\\
%     SYSTEM\_PROMPT<|im\_end|>\\
%     <|im\_start|>user\\
%     Context: \{context\}\\
%     Instruction: \{instruction\}<|im\_end|>\\
%     <|im\_start|>assistant}
%     \end{quote}

The instruction field includes the question prompt or the generic instruction when there are no prompts available. The context  fields includes the following organized political information:
\begin{itemize}[itemsep=2pt, topsep=5pt, parsep=0pt]
    \item \textbf{PARTY:} Political party affiliation (e.g., Conservative)
    \item \textbf{TOPIC:} EuroVoc classification (e.g., TRADE) 
    \item \textbf{SECTION:} Parliamentary debate section
    \item \textbf{POLITICAL ORIENTATION:} Political orientation label (e.g., Right)
    \item \textbf{HOUSE:} House of Commons or House of Lords
\end{itemize}
This setup helps fine-tuned models learn to give responses that match a party’s views, stay on topic, follow parliamentary rules and political views.

%% file: appendix_d.tex
\section{Representative Generated Speeches} \label{appendix:speeches}

We present various speeches generated by the models (baseline and finetuned).
% BASELINE MODEL SPEECHES

\begin{tcolorbox}[speechbox, title={Baseline Model Speech - Gemma}]
\textbf{Political Party:} Scottish National Party \\
\textbf{Topic:} ENERGY --- \textbf{Section:} Domestic Renewable Energy \\
\textbf{House:} House of Commons --- \textbf{Orientation:} Centre-left \\
\textbf{Prompt:} "Address the debate on Domestic Renewable Energy on ENERGY."

\medskip
\hrule
\medskip

\textit{Generated Response:}

Mr Speaker, I rise today to speak about domestic renewable energy as it is 
vital that we ensure our country's future remains secure by reducing reliance 
on foreign imports while also protecting local communities from harmful emissions 
with new clean technologies like solar panels or wind turbines which can be 
installed across all parts of Scotland!
\end{tcolorbox}

\begin{tcolorbox}[speechbox, title={Baseline Model Speech - Llama}]
\textbf{Political Party:} Conservative \\
\textbf{Topic:} INDUSTRY --- \textbf{Section:} Hawk Aircraft: Overseas Promotion \\
\textbf{House:} House of Commons --- \textbf{Orientation:} Centre-right \\
\textbf{Prompt:} "Address the debate on Hawk Aircraft: Overseas Promotion on INDUSTRY."

\medskip
\hrule
\medskip

\textit{Generated Response:}

The government's recent decision to promote exports of our world-renowned aircraft manufacturing capabilities is both commendable and timely. As an advocate for economic growth and international trade, I fully support this initiative. Overseas promotion plays a crucial role in boosting Britain's economy by increasing export revenue. It also helps create jobs domestically while enhancing our nation's reputation as a global leader in innovation and engineering excellence. Hawk Aircraft has long been synonymous with quality craftsmanship, safety standards, and technological advancements that set it apart from its competitors worldwide. By highlighting these strengths during overseas promotional campaigns, we can secure lucrative contracts abroad, ensuring sustained employment opportunities here at home. The current market trends indicate growing demand for advanced fighter jets across various regions globally. With our exceptional track record, skilled workforce, and robust infrastructure supporting research \& development initiatives – now more than ever before - there exists ample scope for us to capitalise upon emerging prospects beyond Europe too! In conclusion then; let me urge my esteemed colleagues within parliament today to rally behind efforts aimed towards fostering increased foreign sales of Hawks through strategic outreach programmes designed specifically tailored according to individual markets' needs thereby contributing significantly toward bolstering overall industrial output alongside creating much needed job openings amidst ongoing challenges faced due COVID19 pandemic.
\end{tcolorbox}

\begin{tcolorbox}[speechbox, title={Baseline Model Speech - Mistral}]
\textbf{Political Party:} Conservative \\
\textbf{Topic:} ENVIRONMENT --- \textbf{Section:} The Situation in the Gulf \\
\textbf{House:} House of Commons --- \textbf{Orientation:} Centre-right \\
\textbf{Prompt:} "Address the debate on The Situation in the Gulf on ENVIRONMENT."

\medskip
\hrule
\medskip

\textit{Generated Response:}

The Conservatives have long supported the EU's position as an independent global player, and we believe that this is particularly important in the context of environmental policy. We recognize the importance of working with our European partners to address issues such as climate change and biodiversity loss, but also acknowledge the need to work closely with countries outside Europe to achieve these goals. This is especially true when it comes to addressing the challenges posed by pollution, deforestation, and other threats to our planet. As such, I would like to take this opportunity to reiterate my support for the EU's commitment to leading the way on environmental protection and call upon all Member States to join us in making sure that the Gulf region remains safe from further damage caused by human activities. We must ensure that any trade agreements between the EU and its trading partners include strong provisions for protecting the environment, including measures to reduce greenhouse gas emissions and preserve natural resources. Additionally, we should continue to encourage dialogue with non-EU states on matters related to sustainable development so that they can benefit from our experience while avoiding potential conflicts over resource allocation or access rights. Finally, I urge my colleagues here today to consider how best we can leverage existing partnerships within both the GCC and Asia Pacific regions towards achieving greater cooperation across borders when tackling shared problems affecting regional ecosystems and economies alike.
\end{tcolorbox}

\begin{tcolorbox}[speechbox, title={Top Scoring Baseline Model Speech - Qwen}]
\textbf{Political Party:} Conservative \\
\textbf{Topic:} POLITICS --- \textbf{Section:} Right-wing Extremism \\
\textbf{House:} House of Commons --- \textbf{Orientation:} Centre-right \\
\textbf{Prompt:} "Address the debate on Right-wing Extremism on POLITICS."

\medskip
\hrule
\medskip

\textit{Generated Response:}

The following is an example of how to write this type of speech: "Mr Speaker, I rise today to address the issue of right-wing extremism within our society. As members of the Conservative Party, we believe that it is important to uphold traditional values and promote social cohesion while respecting diversity. However, there have been concerns raised about some individuals who hold extremist views which run counter to these principles. We must be clear - any form of hate or intolerance has absolutely no place in our country. We cannot allow those with extreme beliefs to divide us based on race, religion or ethnicity. Instead, we need to work together as one nation to build bridges between different communities and foster understanding. In order to combat right-wing extremism, I propose three key actions. Firstly, we need stronger laws against hate crimes so that those responsible can face justice more effectively. Secondly, education plays a vital role; schools must teach young people about tolerance and respect from an early age. Finally, increased engagement by law enforcement agencies will help identify potential threats before they escalate into violence." This speech reflects the centre-right position of the conservative party on the topic of right-wing extremism whilst still maintaining a positive tone towards building unity across society. It also highlights specific solutions proposed by the speaker such as tougher legislation, improved educational programs and greater police involvement.
\end{tcolorbox}

\begin{tcolorbox}[speechbox, title={Baseline Model Speech - Yi}]
\textbf{Political Party:} Labour \\
\textbf{Topic:} AGRI-FOODSTUFFS --- \textbf{Section:} Food Waste \\
\textbf{House:} House of Commons --- \textbf{Orientation:} Centre-left \\
\textbf{Prompt:} "Address the debate on Food Waste on AGRI-FOODSTUFFS."

\medskip
\hrule
\medskip

\textit{Generated Response:}

The EU has an obligation to reduce food waste through sustainable policies that ensure fair prices while also promoting social justice and environmental sustainability. As such, I call upon my fellow parliamentarians from across all parties to support this initiative as it is crucial not only for our planet but also for ensuring equitable access to healthy diets around Europe. This motion must be passed without delay so we can begin implementing these important changes today! Speech generated by AI assistant.
\end{tcolorbox}

% FINETUNED MODEL SPEECHES

\begin{tcolorbox}[speechbox, title={Top Scoring Finetuned Model Speech - Gemma}]
\textbf{Political Party:} Conservative \\
\textbf{Topic:} LAW --- \textbf{Section:} HM Courts and Tribunal Service \\
\textbf{House:} House of Commons --- \textbf{Orientation:} Centre-right \\
\textbf{Prompt:} "Address the debate on HM Courts and Tribunal Service on LAW."

\medskip
\hrule
\medskip

\textit{Generated Response:}

The hon. Gentleman is absolutely right to say that we must do everything possible—and I think everyone across this Chamber agrees with him about that—to ensure victims come forward as witnesses when they can provide evidence against perpetrators who have committed such crimes; but he will also know that there has been an enormous amount of work done over many years by successive Governments since at least the early '90s or later to improve how our justice system deals with vulnerable people so that it does not put them off coming into court. As part of these proposals, which were consulted upon very extensively last year, one thing we wanted was more specialist courts where all those involved could be trained up properly to deal sensitively with vulnerable victims, especially children. There would then be only three types of Crown Court sitting centre around England and Wales, each providing different levels of expertise depending on what type of case was being heard there.
\end{tcolorbox}

\begin{tcolorbox}[speechbox, title={Finetuned Model Speech - Llama}]
\textbf{Political Party:} Conservative \\
\textbf{Topic:} INTERNATIONAL ORGANISATIONS --- \textbf{Section:} UK-EU Trade Policy Options \\
\textbf{House:} House of Commons --- \textbf{Orientation:} Centre-right \\
\textbf{Prompt:} "Address the debate on UK-EU Trade Policy Options on INTERNATIONAL ORGANISATIONS."

\medskip
\hrule
\medskip

\textit{Generated Response:}

I am grateful to my right hon. Friend for his response. As he will be aware, I have long campaigned against our membership of the European Union because we were told that it was not possible to leave without damaging ourselves economically; yet here we are with an economy growing faster than any other major advanced country's. Does this not prove that those who warned us about economic damage if we left were wrong? Will my right hon. Friend assure me that when negotiations start there will be no question but that we must take back control over laws relating to trade policy?
\end{tcolorbox}

\begin{tcolorbox}[speechbox, title={Finetuned Model Speech - Mistral}]
\textbf{Political Party:} Conservative \\
\textbf{Topic:} INTERNATIONAL RELATIONS --- \textbf{Section:} Nuclear Weapons (International Relations Committee Report) \\
\textbf{House:} House of Commons --- \textbf{Orientation:} Centre-right \\
\textbf{Prompt:} "Address the debate on Nuclear Weapons (International Relations Committee Report) on INTERNATIONAL RELATIONS."

\medskip
\hrule
\medskip

\textit{Generated Response:}

2018 is a year of major anniversaries in our history of nuclear weapons. It marks 70 years since atomic bombs were dropped on Japan; it also marks 70 years since NATO was established—I am sure that will be remembered by many Members on both sides of this House; and it also marks 70 years since the UN General Assembly adopted its first resolution calling for an end to nuclear testing. In addition, we mark this year the start of negotiations between President Reagan and Mikhail Gorbachev towards reducing nuclear stockpiles. I commend my noble Friend Lord Howell for his work as Chair of the International Relations Select Committee, which produced the report on nuclear deterrence, "Time to Stop the Clock", earlier this year.
\end{tcolorbox}

\begin{tcolorbox}[speechbox, title={Top Scoring Finetuned Model Speech - Qwen}]
\textbf{Political Party:} Labour \\
\textbf{Topic:} POLITICS --- \textbf{Section:} Hong Kong Arrests Under National Security Law \\
\textbf{House:} House of Lords --- \textbf{Orientation:} Centre-left \\
\textbf{Prompt:} "Address the debate on Hong Kong Arrests Under National Security Law on POLITICS."

\medskip
\hrule
\medskip

\textit{Generated Response:}

My Lords, I am grateful to my noble friend Lord Collins for tabling this Urgent Question today. I associate myself with his comments about human rights lawyer Maya Wang being detained by Chinese authorities last week. It is truly shocking that we have seen more than two dozen people arrested under China's national security law over recent days alone; some were simply trying to exercise their right to free assembly or speak out against the actions of Beijing. This is just one example of how the national security law has been used as an instrument of repression since it came into force late last year.
\end{tcolorbox}

\begin{tcolorbox}[speechbox, title={Top Scoring Finetuned Model Speech - Yi}]
\textbf{Political Party:} Scottish National Party \\
\textbf{Topic:} INTERNATIONAL RELATIONS --- \textbf{Section:} Flooding: EU Solidarity Fund \\
\textbf{House:} House of Commons --- \textbf{Orientation:} Centre-left \\
\textbf{Prompt:} "Address the debate on Flooding: EU Solidarity Fund on INTERNATIONAL RELATIONS."

\medskip
\hrule
\medskip

\textit{Generated Response:}

I thank the Minister for his response to my urgent question earlier today about this issue. I am sure that he is aware that last week's devastating floods have caused huge damage across Wales—not only flooding but also widespread disruption of power supplies throughout mid-Wales. As well as seeking support from our own Government here at home, can we seek solidarity with those affected by these events? Can he update us further on what discussions there will be with other European Governments over how they might help financially or otherwise with such devastation? Will he make representations when it comes time for any applications to be made so that we get maximum financial benefit out of them? Finally, does he agree that while Brexit means leaving Europe physically, not joining together through co-operation makes absolutely zero sense economically, environmentally or socially? This shows why Scotland would rather remain part of Europe than leave altogether.
\end{tcolorbox}

%% file: appendix_e.tex
\section{Statistical Analysis and Detailed Results} \label{app:statistical_tests}

\subsection{Party-Specific Performance}

Table \ref{tab:party_performance} presents detailed performance scores for all UK parliamentary parties across fine-tuned models.

% \input{tables/party_performance}
%\begin{table}[!htbp]
\begin{table*}[!h]
    \centering
    %\small
    \resizebox{\textwidth}{!}{
\begin{tabular}{lrrrrrrr}
\toprule
Party & Gemma 2 9B & Llama 3.1 8B & Mistral 7B v0.3 & Qwen2 7B & YI 6B & Average & Std \\
\midrule
Bishops & 0.609 & 0.624 & 0.550 & 0.664 & 0.562 & 0.602 & 0.042 \\
Conservative & 0.576 & 0.561 & 0.487 & 0.560 & 0.569 & 0.551 & 0.032 \\
Crossbench & 0.599 & 0.623 & 0.453 & 0.574 & 0.592 & 0.568 & 0.060 \\
Democratic Unionist Party & 0.630 & 0.559 & 0.593 & 0.624 & 0.594 & 0.600 & 0.025 \\
Green Party & 0.614 & 0.585 & 0.522 & 0.572 & 0.586 & 0.576 & 0.030 \\
Independent & 0.587 & 0.541 & 0.482 & 0.549 & 0.521 & 0.536 & 0.034 \\
Labour & 0.612 & 0.607 & 0.540 & 0.600 & 0.607 & 0.593 & 0.027 \\
Liberal Democrats & 0.598 & 0.580 & 0.524 & 0.558 & 0.581 & 0.568 & 0.025 \\
Non-Affiliated & 0.620 & 0.554 & 0.436 & 0.526 & 0.633 & 0.554 & 0.071 \\
Plaid Cymru & 0.619 & 0.558 & 0.530 & 0.572 & 0.577 & 0.571 & 0.029 \\
Scottish National Party & 0.597 & 0.622 & 0.543 & 0.596 & 0.614 & 0.594 & 0.028 \\
\bottomrule
\end{tabular}
 }
    \caption{Model Performance by Political Party (Fine-Tuned Models Only)}
    \label{tab:party_performance}
\end{table*} 

\subsection{Metric Validation: Political Discrimination Analysis}

Tables \ref{tab:party_difficulty_rankings} and \ref{tab:orientation_difficulty} present ANOVA results validating the discriminative power of the novel political authenticity metrics.
\begin{table}[!h]

\begin{tabular}{lccc}
\toprule
Rank & Political Party & Difficulty Score & Consistency Score \\
\midrule
1 & Democratic Unionist Party & 0.456355 & 17.207189 \\
2 & Plaid Cymru & 0.454357 & 17.664821 \\
3 & Independent & 0.450960 & 16.126553 \\
4 & Green Party & 0.449039 & 18.088600 \\
5 & Conservative & 0.447174 & 18.570069 \\
6 & Labour & 0.445414 & 17.817318 \\
7 & Liberal Democrats & 0.445108 & 18.165583 \\
8 & Scottish National Party & 0.441675 & 18.100135 \\
9 & Bishops & 0.428796 & 17.796382 \\
10 & Crossbench & 0.425935 & 20.236259 \\
11 & Non-Affiliated & 0.381897 & 19.673312 \\
\bottomrule
\end{tabular}
\caption{Party Difficulty Rankings: Political Parties Ranked by Modeling Difficulty}
\label{tab:party_difficulty_rankings}
\end{table}

\begin{table}[!h]
\begin{tabular}{lccc}
\toprule
Rank & Political Orientation & Difficulty Score & Consistency Score \\
\midrule    
1 & Right & 0.456355 & 17.207189 \\
2 & Centre-left to left & 0.454357 & 17.664821 \\
3 & Left & 0.449039 & 18.088600 \\
4 & Centre-right & 0.447174 & 18.570069 \\
5 & Centre to centre-left & 0.445108 & 18.165583 \\
6 & Centre-left & 0.445095 & 17.800365 \\
7 & Unknown & 0.439036 & 17.085963 \\
\bottomrule
\end{tabular}
\caption{Orientation Difficulty Rankings: Political Orientations Ranked by Modeling Difficulty}
\label{tab:orientation_difficulty}
\end{table}

\subsection{Topic-Specific Performance}

Table \ref{tab:topic_performance} presents model performance across topic domains for fine-tuned models.

% \begin{table}
% \centering
% \small
% \setlength{\tabcolsep}{3pt}
\begin{table*}[!htbp]
    \centering
    %\small
    \resizebox{\textwidth}{!}{
\begin{tabular}{lrrrrrrr}
\toprule
Topic & Gemma 2 9B & Llama 3.1 8B & Mistral 7B v0.3 & Qwen2 7B & Yi 6B & Average & Std \\
\midrule
Agri-Foodstuffs & 0.549 & 0.538 & 0.475 & 0.535 & 0.565 & 0.532 & 0.031 \\
Agriculture, Forestry and Fisheries & 0.587 & 0.572 & 0.491 & 0.579 & 0.568 & 0.559 & 0.035 \\
Business and Competition & 0.606 & 0.566 & 0.528 & 0.577 & 0.582 & 0.572 & 0.026 \\
Economics & 0.626 & 0.621 & 0.553 & 0.615 & 0.636 & 0.610 & 0.029 \\
Education and Communications & 0.581 & 0.575 & 0.492 & 0.546 & 0.565 & 0.552 & 0.032 \\
Employment and Working Conditions & 0.587 & 0.583 & 0.507 & 0.569 & 0.585 & 0.566 & 0.030 \\
Energy & 0.584 & 0.564 & 0.494 & 0.563 & 0.566 & 0.554 & 0.031 \\
Environment & 0.553 & 0.550 & 0.483 & 0.546 & 0.562 & 0.539 & 0.028 \\
European Union & 0.633 & 0.612 & 0.539 & 0.609 & 0.637 & 0.606 & 0.035 \\
Finance & 0.614 & 0.601 & 0.546 & 0.608 & 0.599 & 0.594 & 0.024 \\
Geography & 0.615 & 0.593 & 0.532 & 0.594 & 0.607 & 0.588 & 0.029 \\
Industry & 0.591 & 0.564 & 0.481 & 0.565 & 0.580 & 0.556 & 0.039 \\
International Organisations & 0.583 & 0.577 & 0.490 & 0.575 & 0.586 & 0.562 & 0.036 \\
International Relations & 0.585 & 0.560 & 0.489 & 0.552 & 0.559 & 0.549 & 0.032 \\
Law & 0.592 & 0.580 & 0.505 & 0.575 & 0.579 & 0.566 & 0.031 \\
Politics & 0.617 & 0.605 & 0.549 & 0.597 & 0.592 & 0.592 & 0.023 \\
Production, Technology and Research & 0.562 & 0.561 & 0.482 & 0.555 & 0.563 & 0.545 & 0.031 \\
Science & 0.539 & 0.521 & 0.460 & 0.545 & 0.513 & 0.516 & 0.030 \\
Social Questions & 0.592 & 0.583 & 0.511 & 0.589 & 0.600 & 0.575 & 0.032 \\
Trade & 0.591 & 0.576 & 0.514 & 0.553 & 0.574 & 0.562 & 0.027 \\
Transport & 0.581 & 0.581 & 0.509 & 0.572 & 0.574 & 0.563 & 0.027 \\
\bottomrule
\end{tabular}
}
    \caption{Model Performance by Topic Domain (Party Alignment Scores)}
    \label{tab:topic_performance}
\end{table*} 
%\end{table}
%\input{tables/topic_performance}

\subsection{Cross-Context Stability Analysis}

Table \ref{tab:stability_breakdown} presents cross-context stability scores. Fine-tuned models maintain high consistency across political contexts (91.4-96.2), with Mistral achieving highest overall stability (96.2).

\begin{table}[!h]
\centering
\small
\caption{Cross-Context Stability Analysis by Model and Dimension}
\label{tab:stability_breakdown}
\begin{tabular}{lrrrr}
\toprule
\textbf{Model} & \textbf{Affiliation} & \textbf{Topic} & \textbf{Orientation} & \textbf{Overall} \\
\midrule
Llama 3.1 8B & 92.0 & 97.6 & 95.7 & 95.1 \\
Gemma 2 9B & 88.3 & 95.0 & 91.0 & 91.4 \\
Mistral 7B v0.3 & 94.0 & 97.2 & 97.4 & 96.2 \\
Qwen2 7B & 92.7 & 97.7 & 96.4 & 95.6 \\
Yi 6B & 91.0 & 97.9 & 95.5 & 94.8 \\
\bottomrule
\end{tabular}
\end{table}

\subsection{Statistical Significance Analysis}

Table \ref{tab:ttest_significant_results} presents complete pairwise t-test results comparing baseline and fine-tuned models across all evaluation metrics, including p-values, effect sizes, and significance after Bonferroni correction.

%\input{tables/ttest_results_table}

% \begin{table}[!htbp]
% \centering
% \scriptsize
% \adjustbox{max width=\textwidth}{

\begin{table*}[!ht]
    \centering
    %\small
    \resizebox{\textwidth}{!}{
\begin{tabular}{lllcccc}
\toprule
Test Type & Comparison & Metric & t-statistic & p-value & Cohen's d & Mean Diff \\
\midrule
Fine-Tuning Impact & llama: Baseline vs Fine-tuned & PPL & -42.093 & 0.0000 & -1.134 & -29.5273 \\
Fine-Tuning Impact & llama: Baseline vs Fine-tuned & Dist-N & -28.085 & 0.0000 & -0.757 & -0.0140 \\
Fine-Tuning Impact & llama: Baseline vs Fine-tuned & Self-BLEU & 54.928 & 0.0000 & 1.499 & 0.0119 \\
Fine-Tuning Impact & llama: Baseline vs Fine-tuned & GRUEN & -16.938 & 0.0000 & -0.456 & -0.0546 \\
Fine-Tuning Impact & llama: Baseline vs Fine-tuned & BERTScore & 45.713 & 0.0000 & 1.231 & 0.0150 \\
Fine-Tuning Impact & llama: Baseline vs Fine-tuned & MoverScore & 27.713 & 0.0000 & 0.747 & 0.0048 \\
Fine-Tuning Impact & llama: Baseline vs Fine-tuned & PSA & 23.746 & 0.0000 & 0.648 & 0.0907 \\
Fine-Tuning Impact & llama: Baseline vs Fine-tuned & Party Align & 22.698 & 0.0000 & 0.611 & 0.0739 \\
Fine-Tuning Impact & llama: Baseline vs Fine-tuned & J\_Coh & 16.528 & 0.0000 & 0.445 & 0.8599 \\
Fine-Tuning Impact & llama: Baseline vs Fine-tuned & J\_Conc & 16.016 & 0.0000 & 0.431 & 1.2362 \\
Fine-Tuning Impact & llama: Baseline vs Fine-tuned & J\_Rel & 9.013 & 0.0000 & 0.243 & 0.6684 \\
Fine-Tuning Impact & llama: Baseline vs Fine-tuned & J\_Auth & 23.204 & 0.0000 & 0.625 & 1.7689 \\
Fine-Tuning Impact & llama: Baseline vs Fine-tuned & J\_PolApp & 18.941 & 0.0000 & 0.510 & 1.0925 \\
Fine-Tuning Impact & llama: Baseline vs Fine-tuned & J\_Qual & 9.903 & 0.0000 & 0.267 & 0.6190 \\
Fine-Tuning Impact & gemma: Baseline vs Fine-tuned & PPL & 8.583 & 0.0000 & 0.231 & 12.5989 \\
Fine-Tuning Impact & gemma: Baseline vs Fine-tuned & Dist-N & -4.927 & 0.0000 & -0.133 & -0.0021 \\
Fine-Tuning Impact & gemma: Baseline vs Fine-tuned & Self-BLEU & 18.335 & 0.0000 & 0.501 & 0.0026 \\
Fine-Tuning Impact & gemma: Baseline vs Fine-tuned & GRUEN & -22.653 & 0.0000 & -0.610 & -0.0464 \\
Fine-Tuning Impact & gemma: Baseline vs Fine-tuned & PSA & 14.218 & 0.0000 & 0.388 & 0.0543 \\
Fine-Tuning Impact & gemma: Baseline vs Fine-tuned & Party Align & 15.775 & 0.0000 & 0.425 & 0.0469 \\
Fine-Tuning Impact & gemma: Baseline vs Fine-tuned & J\_Coh & -7.397 & 0.0000 & -0.199 & -0.2794 \\
Fine-Tuning Impact & gemma: Baseline vs Fine-tuned & J\_Auth & 5.658 & 0.0000 & 0.152 & 0.3726 \\
Fine-Tuning Impact & gemma: Baseline vs Fine-tuned & J\_PolApp & 16.112 & 0.0000 & 0.434 & 0.8001 \\
Fine-Tuning Impact & gemma: Baseline vs Fine-tuned & J\_Qual & 9.651 & 0.0000 & 0.260 & 0.5073 \\
Fine-Tuning Impact & mistral: Baseline vs Fine-tuned & PPL & -4.946 & 0.0000 & -0.133 & -2.0489 \\
Fine-Tuning Impact & mistral: Baseline vs Fine-tuned & Dist-N & 5.139 & 0.0000 & 0.138 & 0.0054 \\
Fine-Tuning Impact & mistral: Baseline vs Fine-tuned & Self-BLEU & 33.917 & 0.0000 & 0.926 & 0.0086 \\
Fine-Tuning Impact & mistral: Baseline vs Fine-tuned & BERTScore & 47.147 & 0.0000 & 1.270 & 0.0134 \\
Fine-Tuning Impact & mistral: Baseline vs Fine-tuned & MoverScore & 7.821 & 0.0000 & 0.211 & 0.0015 \\
Fine-Tuning Impact & mistral: Baseline vs Fine-tuned & PSA & 5.168 & 0.0000 & 0.141 & 0.0201 \\
Fine-Tuning Impact & mistral: Baseline vs Fine-tuned & Party Align & -4.586 & 0.0000 & -0.124 & -0.0157 \\
Fine-Tuning Impact & mistral: Baseline vs Fine-tuned & J\_Coh & 22.823 & 0.0000 & 0.615 & 1.3716 \\
Fine-Tuning Impact & mistral: Baseline vs Fine-tuned & J\_Conc & 28.815 & 0.0000 & 0.776 & 2.0922 \\
Fine-Tuning Impact & mistral: Baseline vs Fine-tuned & J\_Rel & 3.850 & 0.0001 & 0.104 & 0.3160 \\
Fine-Tuning Impact & mistral: Baseline vs Fine-tuned & J\_PolApp & 11.594 & 0.0000 & 0.312 & 0.7605 \\
Fine-Tuning Impact & mistral: Baseline vs Fine-tuned & J\_Qual & -6.806 & 0.0000 & -0.183 & -0.4372 \\
Fine-Tuning Impact & qwen: Baseline vs Fine-tuned & PPL & -19.782 & 0.0000 & -0.533 & -8.8869 \\
Fine-Tuning Impact & qwen: Baseline vs Fine-tuned & Self-BLEU & -8.783 & 0.0000 & -0.240 & -0.0029 \\
Fine-Tuning Impact & qwen: Baseline vs Fine-tuned & GRUEN & 11.358 & 0.0000 & 0.306 & 0.0359 \\
Fine-Tuning Impact & qwen: Baseline vs Fine-tuned & BERTScore & 66.712 & 0.0000 & 1.797 & 0.0162 \\
Fine-Tuning Impact & qwen: Baseline vs Fine-tuned & MoverScore & 18.264 & 0.0000 & 0.492 & 0.0029 \\
Fine-Tuning Impact & qwen: Baseline vs Fine-tuned & PSA & 11.771 & 0.0000 & 0.321 & 0.0443 \\
Fine-Tuning Impact & qwen: Baseline vs Fine-tuned & Party Align & 3.680 & 0.0002 & 0.099 & 0.0109 \\
Fine-Tuning Impact & qwen: Baseline vs Fine-tuned & J\_Coh & 4.745 & 0.0000 & 0.128 & 0.1364 \\
Fine-Tuning Impact & qwen: Baseline vs Fine-tuned & J\_Conc & 27.880 & 0.0000 & 0.751 & 1.6796 \\
Fine-Tuning Impact & qwen: Baseline vs Fine-tuned & J\_Rel & -12.751 & 0.0000 & -0.344 & -0.8578 \\
Fine-Tuning Impact & qwen: Baseline vs Fine-tuned & J\_Auth & -11.109 & 0.0000 & -0.299 & -0.8403 \\
Fine-Tuning Impact & qwen: Baseline vs Fine-tuned & J\_PolApp & -3.925 & 0.0001 & -0.106 & -0.1782 \\
Fine-Tuning Impact & qwen: Baseline vs Fine-tuned & J\_Qual & -26.757 & 0.0000 & -0.721 & -1.3538 \\
Fine-Tuning Impact & yi: Baseline vs Fine-tuned & PPL & -34.768 & 0.0000 & -0.937 & -38.5897 \\
Fine-Tuning Impact & yi: Baseline vs Fine-tuned & Dist-N & -11.989 & 0.0000 & -0.323 & -0.0039 \\
Fine-Tuning Impact & yi: Baseline vs Fine-tuned & Self-BLEU & 49.619 & 0.0000 & 1.354 & 0.0095 \\
Fine-Tuning Impact & yi: Baseline vs Fine-tuned & GRUEN & -27.151 & 0.0000 & -0.731 & -0.0503 \\
Fine-Tuning Impact & yi: Baseline vs Fine-tuned & BERTScore & 44.288 & 0.0000 & 1.193 & 0.0180 \\
Fine-Tuning Impact & yi: Baseline vs Fine-tuned & MoverScore & 37.295 & 0.0000 & 1.005 & 0.0062 \\
Fine-Tuning Impact & yi: Baseline vs Fine-tuned & PSA & 38.282 & 0.0000 & 1.045 & 0.1479 \\
Fine-Tuning Impact & yi: Baseline vs Fine-tuned & Party Align & 45.315 & 0.0000 & 1.221 & 0.1569 \\
Fine-Tuning Impact & yi: Baseline vs Fine-tuned & J\_Coh & 25.199 & 0.0000 & 0.679 & 1.3019 \\
Fine-Tuning Impact & yi: Baseline vs Fine-tuned & J\_Conc & 28.626 & 0.0000 & 0.771 & 2.5163 \\
Fine-Tuning Impact & yi: Baseline vs Fine-tuned & J\_Rel & 21.332 & 0.0000 & 0.575 & 1.5595 \\
Fine-Tuning Impact & yi: Baseline vs Fine-tuned & J\_Auth & 47.722 & 0.0000 & 1.286 & 3.1597 \\
Fine-Tuning Impact & yi: Baseline vs Fine-tuned & J\_PolApp & 33.837 & 0.0000 & 0.912 & 1.9474 \\
Fine-Tuning Impact & yi: Baseline vs Fine-tuned & J\_Qual & 41.022 & 0.0000 & 1.105 & 2.3091 \\
\bottomrule
\end{tabular}
}
\caption{Significant Pairwise T-Test Results (Bonferroni Corrected, a=0.05)}
\label{tab:ttest_significant_results}
\end{table*}